\documentclass{article}

\usepackage[final]{corl_2022} 

\usepackage{amsmath}
\usepackage{amsfonts}
\usepackage{mathtools}
\usepackage{amssymb}
\usepackage{wrapfig}
\usepackage[toc,page]{appendix}
\usepackage{booktabs}

\newcommand\Tau{\mathcal{T}}
\DeclareMathOperator*{\argmin}{argmin}

\usepackage{setspace}
\usepackage[subtle]{savetrees}

\usepackage{selectp}

\title{Learning Preconditions of Hybrid Force-Velocity Controllers for Contact-Rich Manipulation}

\author{
  Jacky Liang\\
  Robotics Institute\\
  Carnegie Mellon University \\
  \texttt{jackyliang@cmu.edu} \\
  \And
  Xianyi Cheng \\
  Robotics Institute\\
  Carnegie Mellon University \\
  \texttt{xianyic@cmu.edu} \\
  \AND
  Oliver Kroemer \\
  Robotics Institute\\
  Carnegie Mellon University \\
  \texttt{okroemer@cmu.edu} \\
}

\begin{document}
\maketitle

\begin{abstract}
Robots need to manipulate objects in constrained environments like shelves and cabinets when assisting humans in everyday settings like homes and offices. 
These constraints make manipulation difficult by reducing grasp accessibility, so robots need to use non-prehensile strategies that leverage object-environment contacts to perform manipulation tasks. 
To tackle the challenge of planning and controlling contact-rich behaviors in such settings, this work uses Hybrid Force-Velocity Controllers (HFVCs) as the skill representation and plans skill sequences with learned preconditions.
While HFVCs naturally enable robust and compliant contact-rich behaviors, solvers that synthesize them have traditionally relied on precise object models and closed-loop feedback on object pose, which are difficult to obtain in constrained environments due to occlusions. 
We first relax HFVCs' need for precise models and feedback with our HFVC synthesis framework, then learn a point-cloud-based precondition function to classify where HFVC executions will still be successful despite modeling inaccuracies.
Finally, we use the learned precondition in a search-based task planner to complete contact-rich manipulation tasks in a shelf domain.
Our method achieves a task success rate of $73.2\%$, outperforming the $51.5\%$ achieved by a baseline without the learned precondition.
While the precondition function is trained in simulation, it can also transfer to a real-world setup
without further fine-tuning.
See supplementary materials and videos at~\url{https://sites.google.com/view/constrained-manipulation/}. 
\end{abstract}

\section{Introduction}

Robots operating in human environments, like homes and offices, need to manipulate objects in constrained environments like shelves and cabinets.
These environments introduce challenges in both action and perception.
Environmental constraints reduce grasp accessibility, so robots must use non-prehensile motions that leverage object-environment contacts to perform manipulation tasks.
For example, a book placed in the corner of a shelf has no collision-free antipodal grasps, but a robot can retrieve the book by first pivoting or sliding the book to reveal a grasp.
Environmental constraints also introduce occlusions both before and during robot-object interaction, making precise object modeling and closed-loop vision-based feedback impractical.

In this work, we tackle these challenges by first choosing hybrid force-velocity controllers (HFVCs) that use force control to maintain robot-object contacts and velocity control to achieve the desired motion.
Prior works in HFVCs require accurate object models, object trajectories, known contact modes, and closed-loop feedback of object pose and contacts. 
Our work relaxes these requirements by allowing HFVC synthesis and execution in more realistic settings with incomplete models, partial observations, and without closed-loop object feedback. 
While HFVCs are naturally robust to small modeling errors and collisions, large model mismatches inherent in constrained manipulation may lead to unsuccessful motions. 
To address this, we learn a precondition function that predicts when an HFVC execution will or will not be successful.
The precondition function is a neural network that takes as input segmented point clouds of the scene and the HFVC skill parameters. 
It is trained entirely in simulation, and using point clouds without color information enables easier sim-to-real transfer.
It filters for successful HFVC actions, and we use it in an online search-based task planner to reactively plan sequences of HFVC and Pick-and-Place skills. 
In our experiments, our approach allows a robot to slide, topple, push, and pivot objects as needed to manipulate cuboid and cylindrical objects in an occluded shelf environment.
See Figure~\ref{fig:overview} for an overview of the proposed approach.

The main contributions of our paper are: 
1) A compliant manipulation skill through an HFVC synthesis framework that relaxes previous requirements on object and environment modeling.
This skill can achieve diverse motions like pushing, pivoting, and sliding through generating different parameters.
2) A point-cloud based precondition function for this HFVC skill that predicts if an HFVC execution with the given parameter and current observations would be successful, as some may fail due to inaccurate modeling.
3) Using an search-based planner with the HFVC skill and a Pick-and-Place skill to complete contact-rich manipulation tasks in an occluded shelf domain.
\section{Related Works}

\textbf{Nonprehensile Manipulation.}

Nonprehensile manipulation focuses on controlling contact interactions and planning contact-rich motions.
For control, if we know the object's geometric and dynamics models and relevant contacts, then traditional methods can synthesize controllers that perform behaviors like rolling and slipping~\cite{lynch1996dynamic, lynch1999dynamic}. 
In order to precisely maintain or change the state of contacts, hybrid controllers have been developed based on reasoning about contact modes \cite{doshi2020hybrid, hogan2020reactive, doshi2022manipulation}. 
Compared with traditional methods require accurate models, hybrid and velocity controllers \cite{mason1981compliance} are more robust and have been deployed in industry for decades for tasks such as polishing  
Recently, a hybrid force velocity control (HFVC) method \cite{hou2020manipulation} has been developed for general quasi-static contact-rich motion tasks under uncertainty. 
HFVC is shown to be robust to modeling errors and contact inaccuracies and can perform general constrained manipulation tasks.

Recent works plan multi-step nonprehensile manipulation by directly planning with contact modes with optimization ~\cite{mordatch2012discovery, woodruff2017planning, doshi2020hybrid, aceituno2020global} or sampling based planning methods~\cite{cheng2021contact, cheng2022contact}. 
These methods require accurate models and lack the ability to plan and optimize for longer horizons. 
If skills and transitions are predefined, high-level optimizers like \cite{toussaint2018differentiable} can generate the skill and transition sequence for a longer horizon. 
However, such optimizers are subject to local optima, often not able to find solutions without good initialization. 

Learning approaches can relax modeling assumptions for nonprehensile manipulation.
Some works learn dynamics models~\cite{kutsuzawa2018sequence, pinto2018sample}, while others directly learn policies~\cite{lowrey2018reinforcement, yuan2018rearrangement, sharma2020learning}.
They also make additional assumptions on objects and environments, such as uniform objects and full state information, and they do not plan with different types of nonprehensile strategies.
While there are works on learning-based skill sequencing~\cite{pan2020decision, xu2020deep, simeonov2020long, ichter2020broadly, wang2021learning}, they do not address planning with prehensile skills and different nonprehensile skills in more realistic settings, where skill execution success is unreliable and cannot be directly optimized by the planner.

Compared with the previous works, our method integrates control, planning, and learning into one system that can perform contact-rich tasks in the real world.
We relax modeling assumptions of HFVCs by learning precondition models that predict when skill executions will be successful despite modeling inaccuracies.
We exploit the advantages of HFVCs in a search-based planner --- its ability to represent diverse non-prehensile motions, robustness, and the reduced required computation for planning contact-rich manipulation.

\textbf{Pregrasp Manipulation.}
Another area of related works is pregrasp manipulation, where a robot must perform nonprehensile motions not to directly manipulate an object but to enable future grasps.
A common setting is grasping in clutter, where a target object must be first singulated before it can be grasped.
Singulation is usually done with a pushing policy to maximize downstream grasp, and it can be hardcoded~\cite{danielczuk2018linear}, planned~\cite{dogar2010push}, or learned~\cite{hermans2012guided, zeng2018learning, ni2020learning, correa2019robust}.
In singulation, objects that prevent access to a target object can be pushed away.
In constrained environments, it is the unmovable environment, like tables or shelf walls, that prevent grasping.
Under this context, many works study how to push an object on a table surface over the table edge to expose a grasp.
Some works plan with known object models~\cite{chang2010planning, king2013pregrasp, kappler2012templates}, learned dynamics~\cite{omrvcen2009autonomous}, or explicit surface constraints~\cite{eppner2015planning}, and some directly learn an RL policy~\cite{sun2020learning, zhou2022learning}.
Recent works also proposed online adaptations of object dynamics, like mass and center of mass, to plan for table-edge grasps~\cite{song2020probabilistic, song2020learning}.
While these works show how nonprehensile manipulation can enable downstream grasps, they typically cannot plan with multiple types of nonprehensile behaviors.
Furthermore, they focus on revealing grasps, and not object manipulation in general, which sometimes may not require any grasps.
\section{Method}

\begin{figure}[!t]
    \centering
    \includegraphics[width=0.9\linewidth]{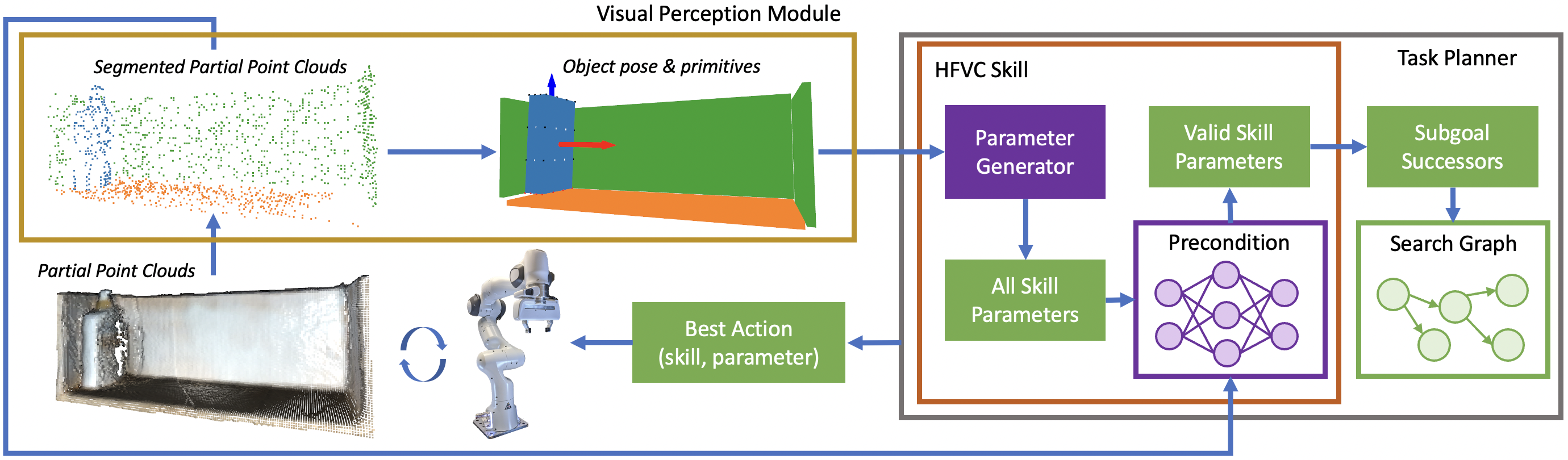}
    \caption{
        \footnotesize
        Overview of our approach that manipulates objects in constrained environments by planning with Hybrid Force-Velocity Controllers (HFVCs).
        Given partial point clouds, we first estimate object-environment segmentations and object and environment geometries.
        These are used to generate HFVC parameters, which are object subgoal poses and robot-object contact points.
        Due to model and feedback inaccuracies, not all generated parameters will lead to successful HFVC executions.
        As such, we learn the HFVC precondition, which predicts skill success from segmented point clouds and skill parameters.
        The planner uses the subgoal poses from the successful parameters to find the current best action, which is a (skill, parameter) tuple.
        The robot then executes this action, and replanning is done as needed.
    }
    \vspace{-10pt}
    \label{fig:overview}
\end{figure}

\textbf{Problem Definition and Assumptions.}
We tackle the problem of moving rigid objects from a start to a goal pose in constrained environments by a robot arm.
The start and goal poses may be in different stable poses, and the object may not have any collision-free antipodal grasps at these poses.
There is only one movable object in the environment.
For actions, the robot arm can perform joint torque control.
For perception, we assume access to an end-effector force-torque sensor and partial point clouds from an RGB-D camera, from which we can estimate segmentation masks and object poses.
For skill parameter generation, we assume object geometries are similar to known geometric primitives, and that we can estimate these primitive shapes and environmental constraints from the segmented point clouds.
Note this assumption does not apply to our learned precondition model and controller synthesis algorithm.
Lastly, we assume object dynamic properties are within a reasonable range that enables nonprehensile manipulation by our robot arm.

\textbf{Approach Overview.}
At the high level, the inputs to our system are partial point clouds that represent the current observation, and the outputs are actions represented as parameterized skills.
Our approach has three main components.
The first is synthesizing HFVCs that allow for compliant execution with inaccurate object models and feedback.
The second is learning the HFVC skill precondition to filter out motions that will likely fail due to modeling mismatches.
The third is using the learned precondition to plan sequences of skills for object manipulation tasks in constrained environments.
The planner uses subgoals from the skill parameters as the skill-level transition model.
The planner replans if the reached state deviates a lot from the subgoal.
See Figure~\ref{fig:overview}.

\textbf{Parameterized Skill Formulation.}
We follow the options formulation of skills~\cite{sutton1999between, konidaris2018skills}.
We denote a parameterized skill as $o$ with parameters $\theta \in \Theta$.
In our work, a parameterized skill $o$ has five elements: a parameter generator that generates both feasible and infeasible $\theta$, a precondition function that classifies skill success given current observations and skill parameters, a controller that performs the skill, a termination condition that tells when the skill should stop, and a skill-level dynamics model that predicts the next state after skill execution terminates.
We assume the parameter generator, controller, and termination conditions are given.
We assume skills have subgoal parameters, which contains information about the next world state if skill execution is successful.
For example, an HFVC skill parameter will contain the desired object pose, and the planner assumes the object will reach the desired pose if 1) that parameter satisfies preconditions and 2) the HFVC commands are computed using that parameter.

\subsection{Hybrid Force Velocity Controller Skill}

The HFVC skill moves an object using a given skill parameter $\theta$, which contains the initial object pose, the desired object pose and the robot-object contact(s). 
Poses are 6D $\operatorname{SE}(3)$.
To achieve the desired object motions, we use an optimization-based solver to output a sequence of hybrid force and velocity commands for the robot to follow.
Below, we explain how parameters are generated for both precondition learning and planning, how HFVC commands are synthesized and executed from the parameter $\theta$, and how we learn the precondition function to classify successful parameters.


\begin{figure}[!t]
    \centering
    \includegraphics[width=1\linewidth]{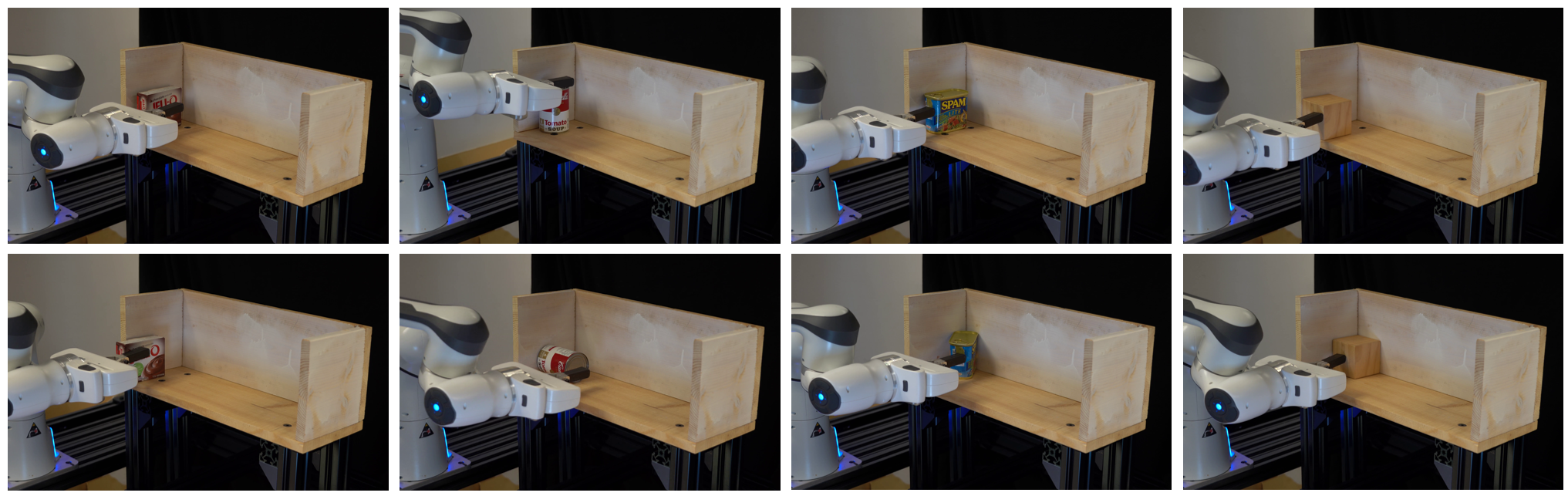}
    \caption{
        \footnotesize
        Our HFVC parameter generator gives diverse contact-rich behaviors.
        Each column is an HFVC skill execution.
        Top: initial states.
        Bottom: final states.
        Left to right: sliding, toppling, pivoting, and pushing.
    }
    \vspace{-5pt}
    \label{fig:hfvc_behaviors}
\end{figure}

\textbf{HFVC Skill Parameter Generation.}
The HFVC skill parameter contains the initial and subgoal object poses and the robot-object contact point(s).
We generate two types of subgoal poses: ones that are in the same stable pose as the current pose, and ones that are in ``neighboring" stable poses (i.e., toppled by $90^\circ$).
Initial robot-object contact points come from evenly spaced points on the surface of object primitives.
We filter out contact points at which the robot would collide with the environment either for the initial or subgoal object poses.
These intended initial contact points should be treated as skill parameters (like ones that parameterize grasps~\cite{sundermeyer2021contact}), as they may not be the actual robot-object contact points achieved during  execution, due to the end-effector geometry and inaccurate pose estimation.
This parameter generation scheme can generate a diverse range of 3D behaviors such as pushing, sliding, pivoting, and toppling.
See Figure~\ref{fig:hfvc_behaviors}.

We assume known geometric primitives (cuboids and cylinders) for parameter generation, but this does not significantly affect the focus of this work.
One reason is that many real-world objects resemble cuboids and cylinders, especially in interacting with constrained environments like shelves and cabinets. 
Another is that this assumption is only made for parameter generation (the precondition directly takes in point clouds), and it can still generate a wide range of parameters with only some satisfying preconditions.
As such, the object primitive assumption does not overly limit the types of behaviors the HFVC skill can achieve, and the parameter generator is useful even if the real object cannot be perfectly represented by the primitives.

\textbf{HFVC Synthesis from Subgoals.}
Our controller synthesis algorithm generates hybrid force-velocity commands $\mathcal{H}$ for the robot to execute.
Initially, the algorithm needs the current object pose $\prescript{W}{}{\Tau}_O \in \operatorname{SE}(3)$, desired object pose $\prescript{W}{}{\Tau}_{sg} \in \operatorname{SE}(3)$, initial estimated object-environment contacts, and initial robot-object contact point. 
An HFVC command $h \in \mathcal{H}$ include the velocity control directions $T_v \in \mathbb{R}^{n_v \times 6}$ and magnitude $\eta_v \in \mathbb{R}^{n_v}$, and the force control direction $T_f \in \mathbb{R}^{n_f \times 6}$ and magnitude $\eta_f \in \mathbb{R}^{n_f}$.
Here, we adopt maximum velocity control where $n_v = 5$ and $n_f = 1$.
During HFVC execution, HFVC commands are computed at a frequency of $20$Hz, and this procedure only needs the current estimated object pose (see next section).

There are three steps in the HFVC synthesis algorithm. 
First, we use a quadratic program (QP) to optimize $T_v$ and $\eta_v$ that moves the object to $\prescript{W}{}{\Tau}_{sg}$ under the contact mode constraints.  
Second, we compute the force control direction $T_f$, which is chosen to be as close as possible to the robot-object contact normal $n_h$ while being as orthogonal as possible to the desired hand velocity direction $v_h$: $T_f = \argmin_{T_f}{(\|T_f v_h\| + \|T_f - n_h\|)}$.
Third, we solve for the force control magnitude $\eta_f$ by trying to maintain a small amount of normal contact forces on every non-separating environment contact under the quasi-dynamic assumption. 
If the robot-object contact normal is parallel to the desired hand velocity direction, we only do velocity control ($n_f = 0, n_v = 6$), which results in a pushing motion.
See details in Appendix.

\begin{wrapfigure}{r}{0.4\textwidth}
  \vspace{-20pt}
  \begin{center}
    \includegraphics[width=0.4\textwidth]{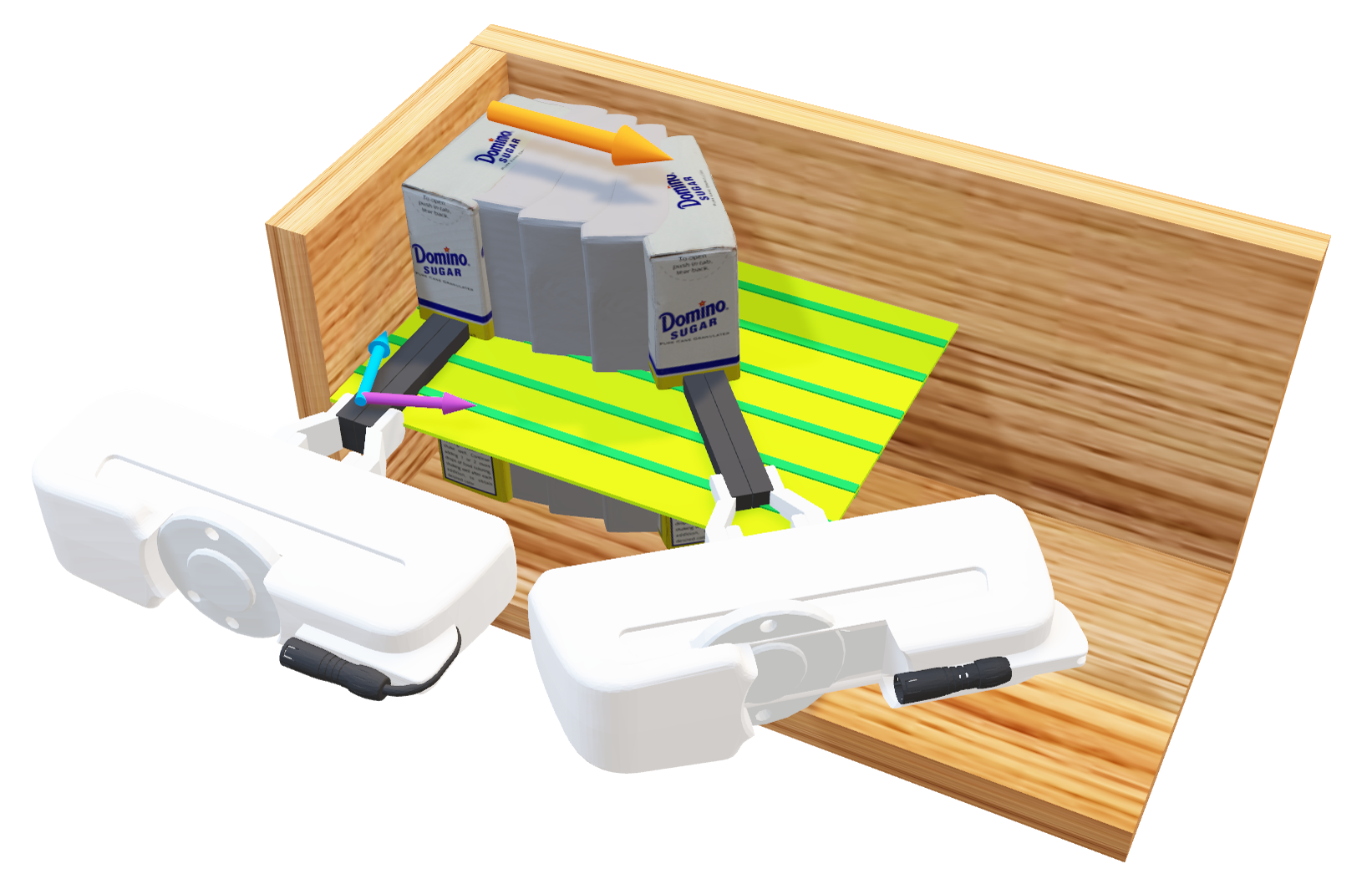}
  \end{center}
  \caption{
    \footnotesize
    HFVC execution with partial information with example pivoting motion (start pose is object on the left).
    Gray objects represent linearly interpolated object poses used by proprioception-based object pose estimation.
    Object translation is exaggerated for visual clarity.
    Blue arrow represents direction of force command, purple is velocity.
    Velocity commands are projected onto the (green) plane containing the interpolated end-effector trajectory.
  }
  \vspace{-10pt}
  \label{fig:hfvc_approx}
\end{wrapfigure}

\textbf{HFVC Execution with Partial Information.}
Computing new HFVC commands during execution requires object pose feedback ($\prescript{W}{}{\Tau}_O$) at about 10Hz, which cannot be directly obtained due to occlusions in constrained environments and delay in perception and state estimation.
As such, we estimate $\prescript{W}{}{\Tau}_O$ from robot proprioception, and we constrain HFVC velocity commands to prevent inaccuracies in this estimation to drastically alter execution behavior.
See Figure~\ref{fig:hfvc_approx}.

We first linearly interpolate a trajectory from the initial pose to the subgoal pose. 
Each point on the trajectory contains an object pose and its corresponding end-effector pose.
To estimate the current object pose from robot proprioception, we assume static robot-object contacts (no slipping) during execution. 
Given the current end-effector pose, we find its closest end-effector pose in the interpolated trajectory. 
The corresponding object pose to the closest end-effector pose is the estimated current object pose.

HFVC is robust to pose estimation errors in the direction of the force controller, which moves the end-effector to maintain the desired robot-object contact.
This is not the case for errors in the direction of the velocity controller, which will keep moving as if the object is still on the interpolated trajectory, leading to execution failure.
To alleviate this issue, we constrain HFVC commands by projecting velocity control onto the plane that contains the interpolated end-effector trajectory, preventing the end-effector from traveling too far from the intended motion. 

\subsection{Learning HFVC Preconditions}

Due to errors in visual perception (e.g. noisy point clouds), real-time feedback (noisy end-effector force-torque sensing, inaccurate object pose estimations), controls (e.g. HFVC solver does not take into account how the low-level controller actually achieves the commanded velocities and forces), and robot-object contacts (they are often non-sticking in practice), HFVC executions are not always successful.
This motivates learning the HFVC skill precondition.
The inputs to the precondition are segmented point clouds and the skill parameter.
The output is whether or not executing the HFVC skill at the given state with the given parameter will be successful.

\textbf{Precondition Success Definition.}
An HFVC skill execution is considered successful if it satisfies three conditions: 1) the object moved more than $1.5$cm or $20^\circ$ after skill execution, 2) the final pose is within $7$cm and $60^\circ$ of the subgoal pose (avoids pose differences that are more than an object's dimension away or in a different stable pose), and 3) the object does not move after the end-effector leaves contact.
Since HFVC executions with model mismatches rarely reach exactly the intended subgoals, having a loose subgoal vicinity requirement allows the planner to execute more HFVC skills and make planning feasible.
These thresholds are specific to a skill and indirectly depend on the object and environment properties.
Tuning them will affect the positive-negative data ratio for training precondition models, but downstream task planning is somewhat robust to these changes, as the planner's prediction threshold for precondition satisfaction can be tuned after the model has been trained.

\textbf{HFVC Data.}
The precondition is trained with HFVC execution data generated in simulation with cuboid and cylindrical-shaped objects from the YCB dataset~\cite{calli2015benchmarking}.
To improve data diversity, we randomize object geometries by sampling non-uniform scales along object principal axes and randomly setting object mass and friction values.
The range for both scaling and dynamics values are chosen so that the resulting object is feasible for manipulation in our shelf domain.
See Figure~\ref{fig:domain} for visualization of the objects used.
We also randomly sample the environment shelf dimensions, as well as the initial object pose and stable pose.
From simulation, we obtain ground truth segmented point clouds and object poses after skill execution, the latter of which is used to compute ground truth skill preconditions.
We use Nvidia IsaacGym, a GPU-accelerated robotics simulator~\cite{liang2018gpu, ig}.

\textbf{Model Architecture.}
The HFVC precondition is a neural network with a PointNet++~\cite{qi2017pointnet++} backbone.
The input point cloud is centered and cropped around the point that is in the center between the initial object position and the subgoal object position.
This improves data efficiency as the network only has to reason about environment points relevant to object-environment interactions.
The features of each point include its 3D coordinate as well as the segmentation label.
For object points, we additionally append the skill parameter as additional features.
These parameter dimensions are set to all $0$'s for the environment points.
The PointNet++ backbone produces embeddings per point.
We take the mean of the point-wise embeddings corresponding to the object points to produce a global object embedding.
This embedding is passed to a multi-layer perceptron (MLP) to produce the final precondition prediction. 
The entire network is trained with a binary cross-entropy loss.

\subsection{Search-Based Task Planning}
Once the precondition is learned, we can use it along with the skill parameter generator to plan for tasks.
We perform task planning in a search-based manner on a directed graph, where each node of the graph corresponds to a planning state, and each directed edge corresponds to a (skill, parameter) whose execution from the source state would result in the next state.
In our domain, a task is specified by an initial and a goal object pose, and the planning state is the 6D object pose.
For parameter generation, if the goal pose is close enough to the current pose, then it is included in the list of subgoal poses, so the planner can find plans that take the object directly to the goal pose.
The output of the planner is a sequence of parameterized skills to be executed from the initial state.

We interleave planning graph construction with graph search, and search is performed in a best-first manner, similar to~\cite{liang2021search} which uses A* to perform task planning.
However, it is difficult to efficiently perform A*-style optimal planning in our domain.
This is due to several factors --- 1) inaccurate transition models because we directly use subgoals as the next state, 2) expensive edge evaluation (node generation requires storing new point clouds for future precondition inference), and 3) high branching factor with many possible skill parameters at a given state.
As such, we instead use Real-Time A*~\cite{korf1990real} with an inadmissible heuristic, where A* is performed until a search budget is exhausted or when a path to the goal is found.
Then, the robot executes the first (skill, parameter) tuple of the path that reaches the best leaf node found so far.
After skill execution, if the observed next state is not close enough to the expected next state, or we have reached the end of the current plan, we will replan again.
\section{Experiments}
\begin{figure}[!t]
    \centering
    \includegraphics[width=0.29\linewidth]{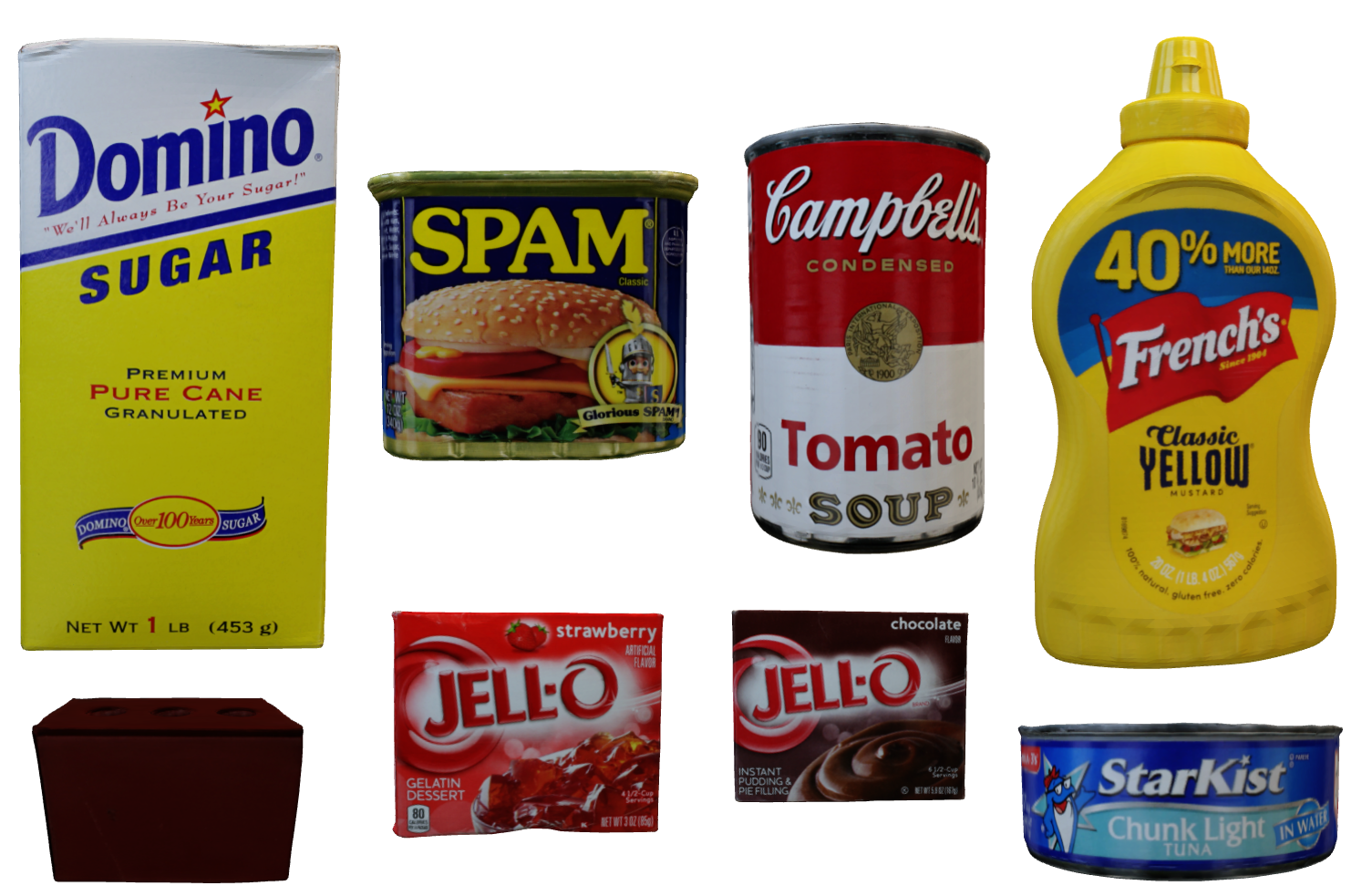}
    \includegraphics[width=0.2\linewidth]{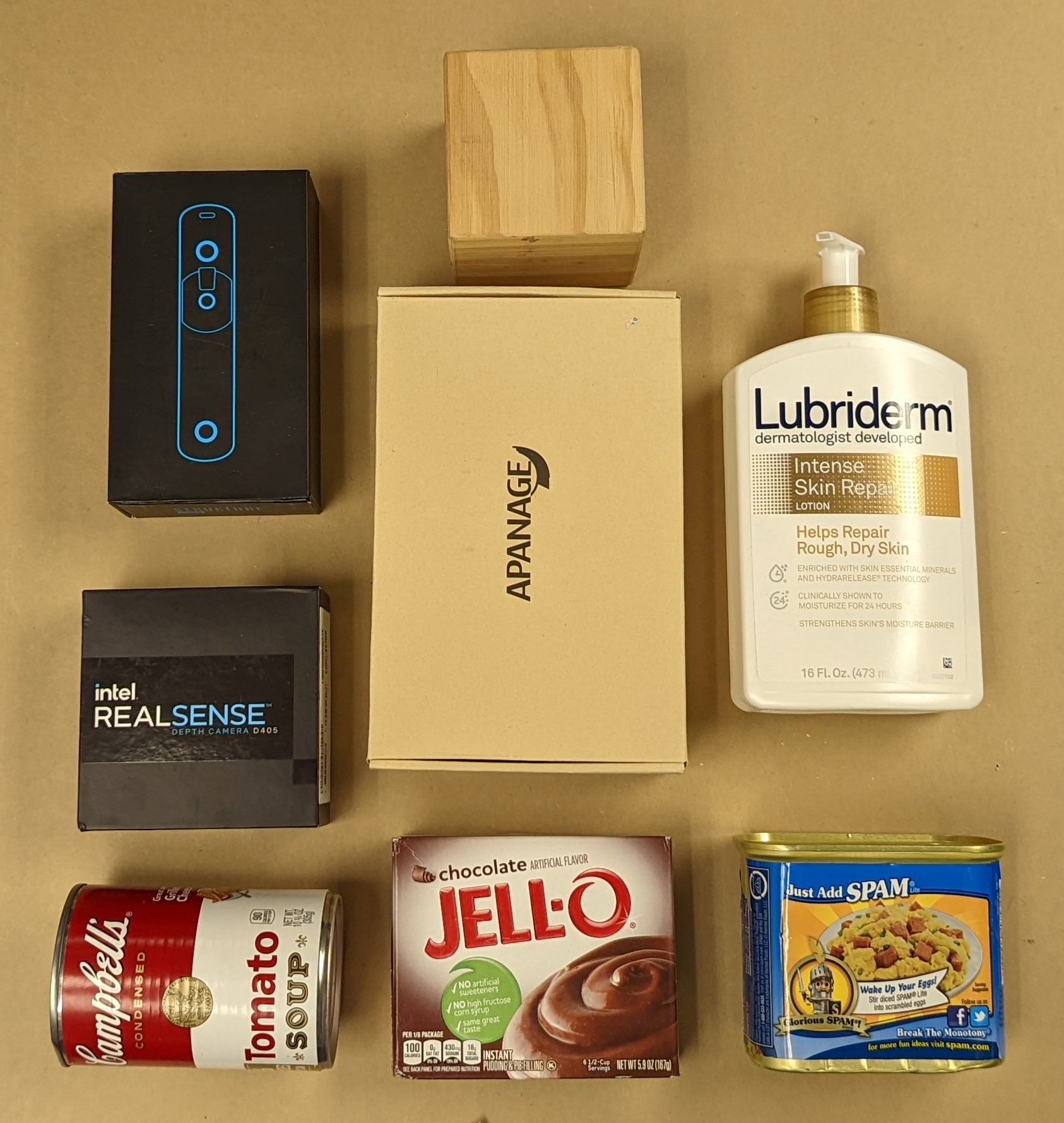}
    \includegraphics[width=0.4\linewidth]{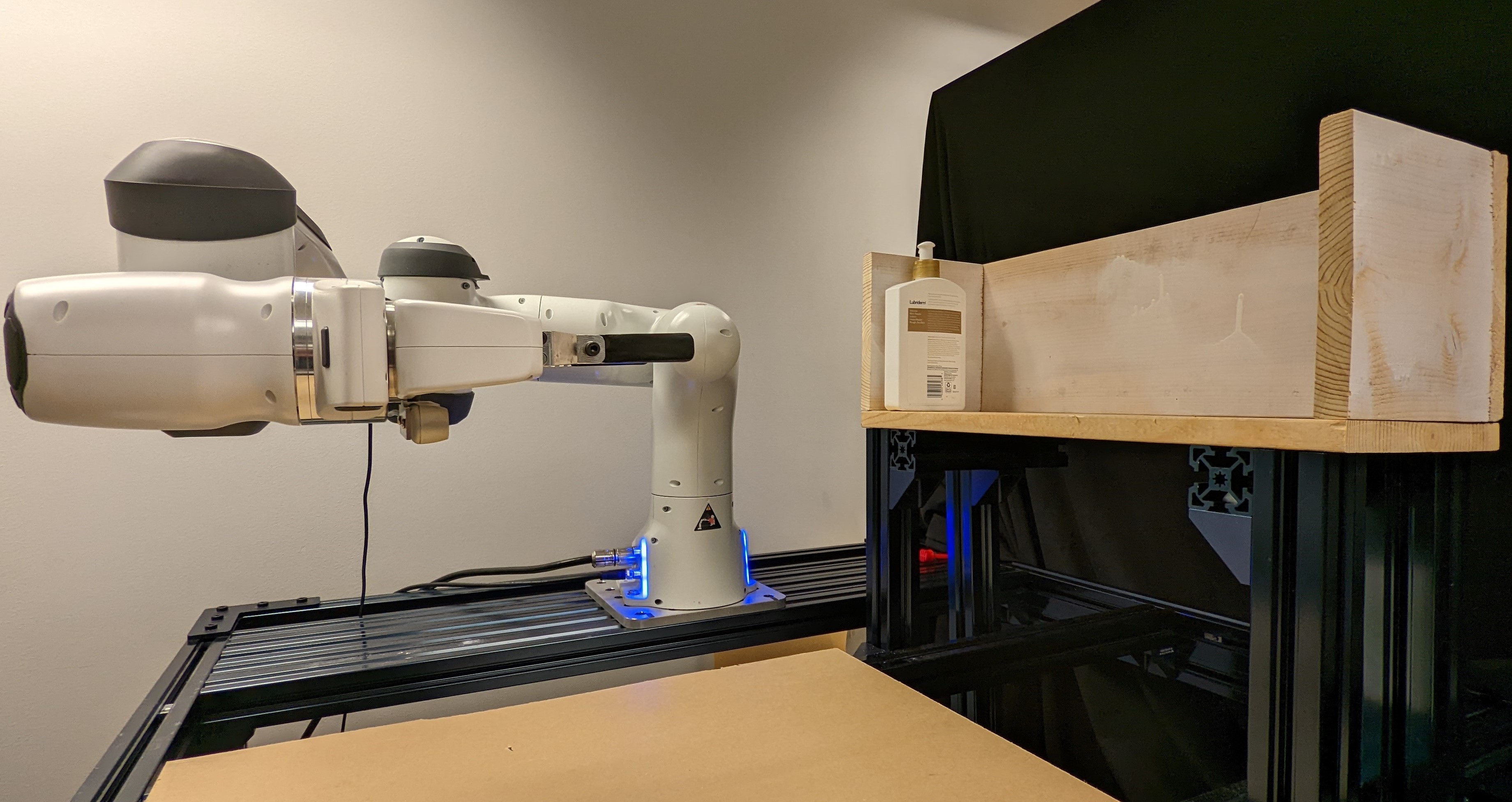}
    \caption{
        \footnotesize
        Object-in-Shelf Task Domain.
        Left: cuboid and cylindrical YCB objects used in simulation for training the HFVC precondition.
        Middle: objects used in real-world experiments.
        Right: real-world setup.
    }
    \vspace{-15pt}
    \label{fig:domain}
\end{figure}

Our experiments focus on evaluating our approach in a shelf domain, where a robot needs to move an object from a start pose to a goal pose.
We perform three types of evaluations.
The first gauges how much proprioception-based object pose estimation and velocity command projections improve HFVC execution success.
The second is about HFVC skill precondition training performance.
The third is on the overall task performance by running the planner with the learned HFVC skill precondition.
A key comparison we make is between planning with and without the learned precondition function.
Lastly, we demonstrate our planner and learned precondition on a real-world robot setup without further model fine-tuning.

\textbf{Task Domain.}
Our task domain consists of a 7 DoF Franka Panda robot arm, a rectangular shelf, and a set of objects for the robot to manipulate.
Each task instance is specified by a different object geometry, shelf dimensions, shelf pose, initial object pose, and goal object pose.
The robot successfully completes the task if it executes a sequence of (skill, parameter) tuples that brings the object from the initial pose to the goal pose, with a goal threshold of $1.5$cm for translation and $10^\circ$ for rotation.
In addition to the HFVC skill, the planner also has access to a Pick-and-Place skill that can move objects via grasps if grasps are available.

\subsection{HFVC Execution with Partial Information.}

\begin{table}[]
\centering
\begin{tabular}{lccc}
\toprule
              & Skill Success     & SG-ADD (cm)             & S-SG-ADD (cm) \\ 
\midrule
Ours          & $\mathbf{52.3\%}$ & $\mathbf{6.9 \pm 8.1}$  & $\mathbf{3.4 \pm 1.9}$         \\
No-Feedback   & $38.6\%$          & $10.2 \pm 10.3$         & $3.9 \pm 1.9$ \\
No-Constraint & $46.0\%$          & $7.6 \pm 8.1$           & $3.6 \pm 2.1$ \\
No-Both     & $38.3\%$          & $10.3 \pm 10.3$         & $3.8 \pm 1.9$ \\
\bottomrule
\end{tabular}
\caption{
    \footnotesize
    Skill execution evaluations for our approach vs. ablations that do not use proprioception-based object pose estimation and planar velocity constraints.
    We report skill success rate and error distance between the subgoal pose and the actual reached pose over all executions (SG-ADD) and over just the successful ones (S-SG-ADD).
    Numbers after $\pm$ are standard deviations.
}
\vspace{-10pt}
\label{tab:hfvc}
\end{table}

We first demonstrate the value of estimating object pose from proprioception and applying planar constraints on velocity commands.
One ablation is running HFVC with ``open-loop" pose estimation, where the current object pose is indexed from the interpolated trajectory with a time-based index (No-Feedback).
This assumes the object is following the interpolated trajectory at a fixed speed.
Another is running HFVC with proprioceptive feedback but without the velocity planar constraints (No-Constraint).
The third variant is running HFVC without both of these modifications (No-Both).
We report the execution success rate, the object's average discrepancy distance (ADD) between achieved and subgoal poses for all executions (SG-ADD) and only for the successful executions (S-SG-ADD).
A total of $7.5$k skill executions per method were used in these evaluations.
See results in Table~\ref{tab:hfvc}.
Our approach achieves $52.3\%$ success rate, higher than the ablations, with proprioception-based pose feedback ($38.6\%$) making the most difference.
These results demonstrate that both proprioceptive object pose feedback and planar velocity constraints can improve HFVC skill executions.

\subsection{Precondition Learning}

\begin{table}[!t]
\centering
\begin{tabular}{lcccc}
\toprule
          & Ours & Full-PC   & GT-Primitive & No-Params \\
\midrule
Accuracy  & $79\%$      & $76\%$ & $\mathbf{85\%}$   & $65\%$          \\
Precision & $78\%$      & $73\%$ & $\mathbf{85\%}$   & $60\%$             \\
Recall    & $77\%$      & $79\%$ & $\mathbf{83\%}$   & $79\%$             \\
\bottomrule
\end{tabular}
\vspace{3pt}
\caption{
    \footnotesize
    Precondition training results with different input representations.
}
\vspace{-15pt}
\label{tab:training_results}
\end{table}

We generate the HFVC execution dataset with 6 cuboid-shaped and 2 cylindrical-shaped objects from the YCB dataset (see Figure~\ref{fig:domain}).
Train-test split is done across object scaling factors, with the test set having the smallest and biggest scales.
To see how the network would perform with full, instead of partial observations, we include one ablation that uses full point clouds (Full-PC) and another that uses vertices of ground-truth object primitive meshes (GT-Primitive).
We also train a variant that does not use skill parameter features (No-Params).
See Table~\ref{tab:training_results}.
Our method using partial point clouds is on par with Full-PC, and GT-Primitive performs the best.
However, we cannot directly use GT-Primitive in planning due to having access to only partial point clouds.
Instead, we include a comparison by using estimated object geometries in the task planning experiments below.
No-Params performs the worst as expected, but it is still able to improve over random guessing due to biases in our collected data.

\subsection{Task Planning Experiments}

\begin{table}[!t]
\centering
\begin{tabular}{llllll}
\toprule
              & Ours              & Est-Primitive        & No-PC             & Only-Pick-Place & No-Replan \\ 
\midrule
Plan Success  & $\mathbf{73.2}\%$ & $61.1\% $       & $51.5\%$          & $27.4\%$        & $28.0\%$ \\
Plan Time (s) & $43.1 \pm 23.0$   & $44.5 \pm 26.5$ & $36.5 \pm 20.3 $  & $6.7 \pm 11.6$  & $22.4 \pm 13.1$           \\
Plan Length   & $3.5 \pm 1.3$     & $3.2 \pm 1.5$   & $3.6 \pm 1.5$     & $1.6 \pm 0.8$   & $1.9 \pm 1.0$ \\
\bottomrule
\end{tabular}
\vspace{3pt}
\caption{
    \footnotesize
    Task performance of our approach using partial point clouds vs. using estimated object primitives (Est-Primitive), not using learned preconditions (No-PC), only using Pick-and-Place (Only-Pick-Place), and not doing replanning (No-Replan).
    Plan time includes replanning time.
    Numbers after $\pm$ are standard deviations.
    Plan time and length statistics are computed only over successful trials.
}
\vspace{-15pt}
\label{tab:planning_results}
\end{table}

We evaluate task planning performance across several ablations in simulation.
For each method, we run trials across the $8$ YCB objects and $8$ task scenarios, with $5$ trials per object and scenario pair, resulting in a total of $240$ trials per method.
Each trial samples different initial and goal poses, and object geometries used in task evaluation are not in the training set.
A task scenario specifies whether or not the initial object pose and goal pose are close to the shelf wall ($4$ variants) and whether they have the same stable pose ($2$ variants), for a total of $8$ scenarios.
We report the overall task success rate, average planning time and plan length for successful trials.

The first ablation is on using vertices from estimated object primitives (Est-Primitive) as the input to the precondition, instead of partial point clouds.
The second is planning without the learned precondition, so the planner treats all generated parameters as feasible (No-PC).
The third evaluates the usefulness of the HFVC skill in our domain by planning only with the Pick-and-Place skill (Only-Pick-Place).
The last ablation does not replan and executes the entire found path from the initial state without feedback (No-Replan).
For this method, we double the planning budget to $60$s, so the planner is more likely to find a path all the way to the goal state.

See Table~\ref{tab:planning_results}.
The proposed approach achieves a success rate of $74.7\%$, higher than Est-Primitive ($61.1\%$) and No-PC ($48.9\%$), and much higher than Only-Pick-Place ($22.9\%$) and No-Replan ($28.8\%$).
The drop in performance of Est-Primitive shows that while using ground truth primitives gives better precondition predictions, this improvement does not apply when primitives must be estimated from partial point clouds.
The other ablations show the importance of learning HFVC preconditions, using HFVC skills in constrained environments, and replanning to compensate for inaccurate subgoal transition models.

\textbf{Real-world Demonstration.}
Lastly, we demonstrate our planner with the learned precondition can operate in the real world.
We use FrankaPy~\cite{zhang2020modular} to control the real Franka arm at $100$Hz, similar to simulation.
While the learned precondition can operate on real-world point clouds without further training, we had to tune low-level controller gains to reproduce similar contact-rich motions on the real robot.
As with simulation, our planner is able to find plans that include a variety of contact-rich behaviors, like pushing, sliding, toppling, and pivoting, to manipulate objects on the shelf.
Please see supplementary materials for more details and real-world videos.

\textbf{Failure Modes and Limitations.}
The most common failure mode is precondition errors that lead the planner into finding infeasible plans or not finding a plan when there is one.
This may be addressed by further improving precondition performance with more data, or enforcing domain-specific invariances.
In addition, the object primitive assumption does prevent the parameter generator from supporting more complex objects, and this may be resolved by using learned parameter generator and visual perception module.
For execution, our planner does not perform online precondition adaptation for unexpected object dynamics; doing so may improve task performance for objects with out-of-distribution dynamics parameters.
Lastly, our method assumes only one movable object in the scene.
Manipulating multiple objects may require learned perception systems and skill-level dynamics models that work in clutter.


\section{Conclusion}
HFVCs can naturally express Contact-rich manipulation behaviors.
However, their reliance on precise models and closed-loop object feedback have prevented their use in cases where such information cannot be easily obtained.
Our work 1) modifies HFVCs to not rely on privileged information and 2) learns where HFVCs are successful despite inaccurate models so 3) a planner can plan sequences of HFVC and Pick-and-Place skills to do contact-rich tasks in constrained environments.


\clearpage


\bibliography{citations}  

\clearpage
\begin{appendices}
\normalsize

\section{Visual Perception Module}

\begin{figure}[!h]
    \centering
    \includegraphics[width=\linewidth]{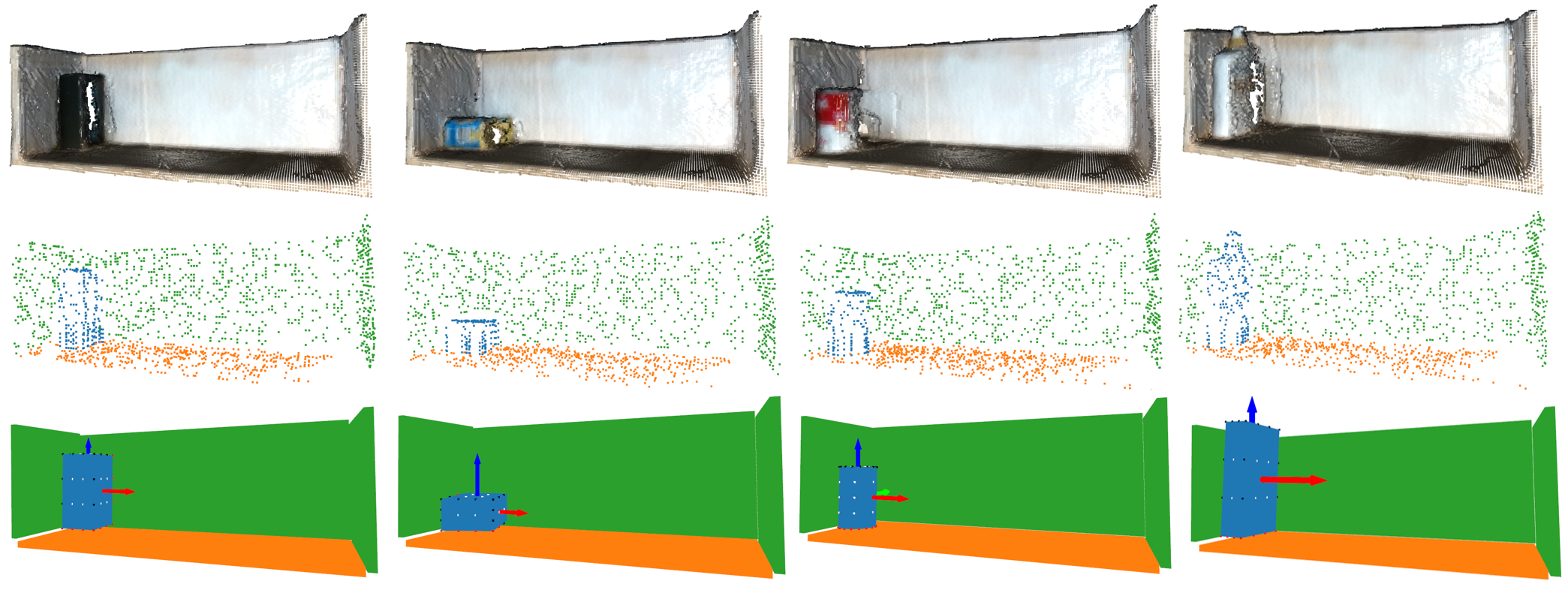}
    \caption{
        \footnotesize
        Additional example outputs of our visual perception module on real-world observations.
        Top row are visualizations of TSDF volumes fused from 6 RGBD images taken around the shelf.
        Note that these are partial observations, and there are missing data in the point clouds.
        Middle row are visualizations of segmented points.
        There are only two segmentation classes the precondition function uses --- object and shelf.
        We color the shelf's bottom points in visualization for clarity.
        The bottom row are estimated geometries for both the objects and the shelf, as well as the estimated object pose.
        Black points on the object surface are potential contact points that are not in contact; red points are contact points that are in contact.
        White points are potential initial robot-object contact points used for parameter generation.
    }
    \label{fig:real_vstates}
\end{figure}

Here we provide additional details on the visual perception module.
It segments object and environment classes from point clouds, and then it extracts object primitive geometries, object pose, and environmental planar constraints from the segmented point clouds.
In our domain that has a rectangular shelf, environment points are segmented out by fitting an oriented bounding box to the entire scene.
Then, we fit the corresponding object primitive geometry to the remaining object points, the centroid and principal axes of which form the object pose.
We initially planned to train neural networks to perform these vision tasks, but we found that our existing pipeline (using Open3D~\cite{Zhou2018} and Trimesh~\cite{trimesh}) was already sufficiently robust for our domain.

For obtaining environment information, finding environmental planar constraints and performing object-environment segmentation can be done together.
We assume the environment has a cuboid shape (e.g. a shelf), so finding points that belong to the environment can be done by first fitting an oriented bounding box and then labeling all points close to the surface of that bounding box as environment points.

For finding the object bounding box and pose of cuboids, we first fit an oriented bounding box to the segmented object points.
This is done by clustering the object normals and treating the cluster centers with the 2 highest counts as the principal directions for the oriented bounding box.
The dimensions of the bounding box can be directly computed by projecting object points onto these directions and computing the corresponding extents.
These directions, along with the centroid of the oriented bounding box, are used to form the object pose.

For fitting cylinder primitives, we first fit an oriented bounding box on the partial point cloud, then use the bounding box's rotation axes as candidates for the cylinder's primary axis (about which there is rotational symmetry). 
For each candidate axis we rotate the partial point cloud around it by 90 degrees and fit another oriented bounding box on both the original and the rotated points. 
The candidate axis that corresponds to the new bounding box with the smallest volume is chosen as the primary rotation axis, and we extract the radius and pose of the cylinder from the dimensions and pose of this bounding box. 

While the visual perception module was sufficient for our experiments, it does have limitations and failure modes that hinder more general use.
Specifically, fitting object geometry primitives to noisy point cloud data is not as reliable in the real world as it is in simulation.
Incorrectly estimated object shapes and poses, leading to infeasible skill parameters, is the most common cause of real-world skill execution failures.
The visual perception module can be improved by using learned models, and this may also allow it to generalize to a broader range of object geometries.
We did not pursue furthering the capabilities of the visual perception module to focus time and engineering effort toward the contributions of this paper, but we are confident that our method could be extended to more complex object geometries as the neither precondition model nor HFVC synthesis rely on these geometric primitives (the precondition model uses point clouds, and HFVC synthesis uses sparse contact points which can be inferred by learned methods).

\section{HFVC Skill}

\subsection{HFVC Parameter Generation}

\begin{figure}[!t]
    \centering
    \includegraphics[width=\linewidth]{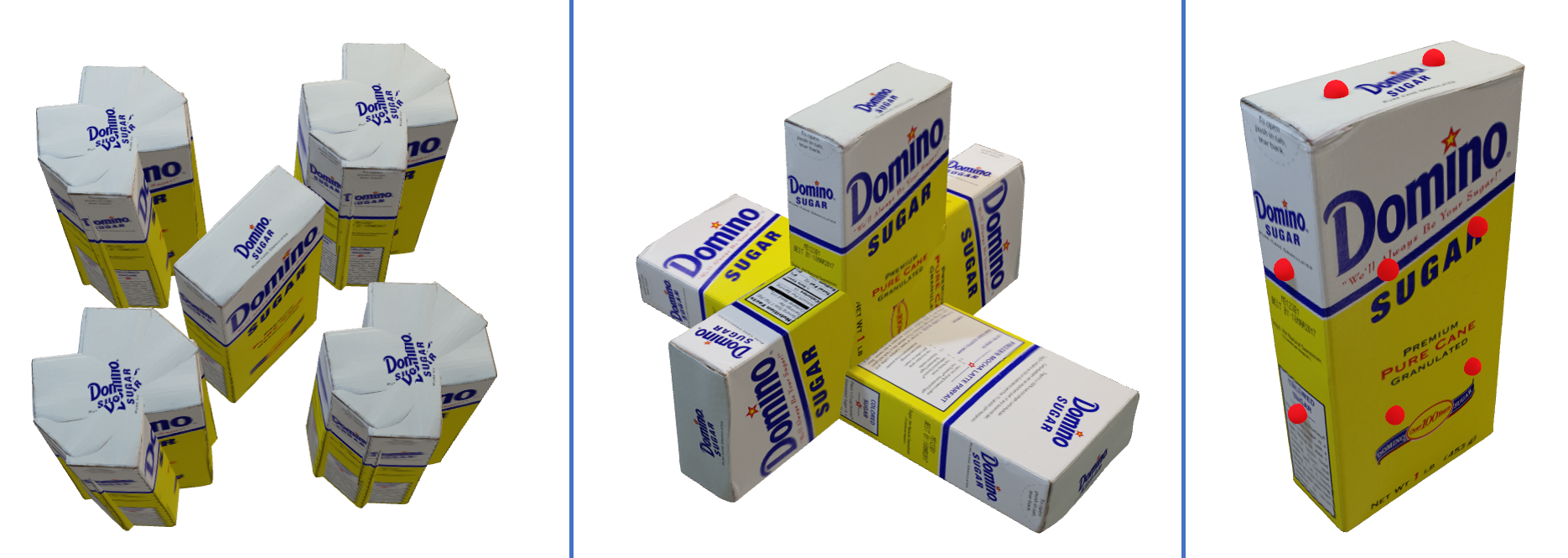}
    \caption{
        \footnotesize
        Subgoal generation. 
        Left: planar subgoal poses --- combination of $4$ translations and $3$ rotations (middle object is the original pose). 
        Translations have been exaggerated for visual clarity.
        Middle: subgoals that are in neighboring stable poses (middle object is original pose).
        Right: Red dots are initial robot-object contact points come from a grid on the object surface.
    }
    \label{fig:subgoals}
\end{figure}

Here we provide more details on HFVC skill parameter generation.
See Figure~\ref{fig:subgoals}
Recall there are two components of the generated parameter --- the desired object subgoal pose and a robot-object contact point.

There are two types of subgoals we generate --- ones in the same stable pose, and ones that are in neighboring stable poses (rotate by $90^\circ$).
For the subgoals in the same stable pose, we form $4$ delta translations along the positive and negative principal directions of the environment bounding box.
This allows us to sample subgoals that move the object along the walls of the environment.
The translation magnitudes are the object bounding box dimensions in the corresponding directions.
This allows the translation magnitudes to automatically scale with object size.
In addition, we form $3$ planar rotations, with degrees $[-30^\circ, 0^\circ, 30^\circ]$ that are applied to each of the $4$ delta translations.
For subgoal poses in different stable poses, we obtain ``neighboring" stable poses that are $90^\circ$ apart by rotating the current object pose around the object bounding box's principal axes.
We do not apply additional planar rotations to these subgoal poses.
The planar rotations and different stable poses allow the HFVC skill to achieve complex behaviors, like pivoting and toppling.
There are at most $4 \times 3 + 4 = 16$ subgoal poses generated, and we filter out the ones that intersect with the environment.
In the context of planning, these subgoal poses can be thought of as motion primitives that are object-centric instead of robot-centric.

For cylinders, when they are vertical, we generate neighboring stable poses by applying 90 degree rotations in the 2 axes that are not the axis of symmetry. 
This looks similar to the cuboid stable pose figure in the Appendix. 
When cylinders are not vertical, only 1 axis would give a different neighboring stable pose (horizontal -> vertical), so we only use that one. 
There are infinite pairs of axes that are orthogonal to the axis of symmetry - we pick the pair that's aligned with the shelf walls, so these subgoals can result in motions that pivot against shelf walls. 

For robot-object contact points, we compute them from a evenly spaced grid on the object primitive surface.
This gives $24$ points for cuboids and $46$ for cylinders.

During execution, the robot approaches the given object contact point, and it proceeds to executing the HFVC after a contact has been detected.
The achieved robot-object contact point is most likely different from the contact point in the parameter, because the object bounding box may not accurately capture the underlying object geometry.
Sometimes, this mismatch leads to HFVC execution failures, and the precondition function would learn to classify these parameters as such, so the planner can avoid such failures during task execution.

\subsection{HFVC Synthesis}

Our HFVC synthesis algorithm outputs the hybrid force velocity control commands in the 6D end effector task space. 
An HFVC command $h \in \mathcal{H}$ include the velocity control directions $T_v \in \mathbb{R}^{n_v \times 6}$ and magnitude $\eta_v \in \mathbb{R}^{n_v}$, and the force control direction $T_f \in \mathbb{R}^{n_f \times 6}$ and magnitude $\eta_f \in \mathbb{R}^{n_f}$. 
While which direction to perform force or velocity control can be arbitrarily chosen, here we adopt maximum velocity control where $n_v = 5$ and $n_f = 1$.

During the controller construction, our algorithm take as inputs the start(current) object pose $\prescript{W}{}{\Tau}_O$, desired object pose $\prescript{W}{}{\Tau}_{sg}$, partially observed object environment contacts, and the robot object contact(s). 
The algorithm first reason about the 3D environment contact mode of the desired motion (i.e. which contacts are separating and which contacts are sliding), by comparing the environment contact locations at the current object pose and desired object poses. 
Our controller will maintain this environment contact mode during execution. 
For robot object contact(s), we assume they are either a sticking point contact or a sticking small patch contact with the object (can be approximated by over three point contacts). 
Now we get the contact mode of all the contacts in the system --- environment contacts could be sticking, sliding or separating. Robot contacts are always sticking. 
And we fix this contact mode. 
A fixed contact mode enables us to safely use continuous constrained dynamics in our following computations. More importantly, by choosing to maintain a contact mode, we don't need to perform contact detection and estimation (which is expensive and slow) during execution. 

During the execution, at each timestep, the controller first update the current object pose estimation $\prescript{W}{}{\Tau}_O$. 
Assuming the object-environment contacts are fixed in the world frame (not moving), we update the environment contact locations and normals in the object frame. 
There are three steps in computing the hybrid force velocity control parameters:
\begin{enumerate}
    \item Compute velocity control directions and magnitudes
    \item Compute force control directions
    \item Compute force control magnitudes
\end{enumerate}
In the first step, we compute the desired contact mode constrained hand velocity $v_h \in \mathbb{R}^6 $ (in body twist form). 
Velocity command $v_h$ is solved with a quadratic programming with which the cost is to get the constrained object velocity $v_o$ as close as possible to the desired object velocity $v_o^{\mathrm{des}}$. $v_o^{\mathrm{des}}$ is computed by the desired object pose and current object pose from using the first order Euler integration.
\begin{equation}
\label{eqn:optvel}
\begin{array}{ll}
     & \min\limits_{v_o, v_h} \| v_o^{\mathrm{des}} - v_o \|_2^2\\
    \text{s.t.} & N \begin{bmatrix} v_o \\ v_h \end{bmatrix} = 0 \text{ (contact mode velocity constraints)} \\
     &  \prescript{W}{}{\Tau}_{sg} =  \prescript{W}{}{\Tau}_O  \cdot \exp{(v_o^{\mathrm{des}})} \text{ (first order Euler integration of object pose)} 
\end{array}
\end{equation}
We scale this hand velocity to be no larger than some threshold. 

In the second step, we determine the force control direction. 
The force control direction is chosen to be as close as possible to the robot object contact normal direction $n_h$ while being as orthogonal as possible to the desired hand velocity direction $v_h$ by having $T_f = \argmin_{T_f}{(\|T_f v_h\| + \|T_f - n_h\|)}$. 
If the robot-object contact normal is parallel to the desired hand velocity direction, we only do velocity control by letting $n_f = 0, n_v = 6$ (this often results in a pushing motion).

In the third step, we solve for the force control magnitude by trying to maintain a small amount of normal contact forces on every non-separating environment contact under quasi-dynamic assumption. 
Denote the contact forces as $\lambda = \begin{bmatrix} \lambda_e \\ \lambda_h \end{bmatrix}$, where $\lambda_e$ is the object-environment contact forces and $\lambda_h$ is the object-hand contact forces.
Denote $\eta = \begin{bmatrix} \eta_f \\ \eta_v \end{bmatrix}$, where $\eta_f$ is the force control magnitude, and $\eta_v$ are the reaction forces of the hand in the velocity control directions. 
Here, ${J_{\lambda}^o}^T$ is the contact jacobian that transforms all contact forces in the wrench space in the object body frame, and ${J_{\lambda_h}^h}^T$ is the contact jacobian than transforms hand contact forces in the wrench space of the hand body frame. 
Matrix $T$ describe the directions of force and velocity control axes. 
Let $M_o$ be the object inertia and $a$ be the quasi-dynamic object acceleration. 
The optimization problem can be written as follows:
\begin{equation}
\label{eqn:contact-forces}
\begin{array}{ll}
     & \min\limits_{\lambda, \eta} \| \lambda - \lambda^{\mathrm{des}} \|_2^2\\
    \text{s.t.} & FC(\lambda) \leq 0 \text{ (friction cone constraints)} \\
     &  {J_{\lambda}^o}^T \lambda + F_{external} = M_o a \text{ (quasi-dynamic balance on the object)} \\
     &  a \in PCC_{v_o} \text{ (object acceleration constraint)} \\
     &  {J_{\lambda_h}^h}^T \lambda_h + T^{-1} \eta = 0  \text{ (force balance on the hand)}
       
\end{array}
\end{equation}
Friction cone constraints ensures all the contact forces satisfy the Coulomb Friction law. 
The object quasi-dynamic balance constraints allowing object to have acceleration $a$ which has the direction in a small polyhedral convex cone aligned with desired object velocity $v_o$, while ignoring Coriolis forces. 
The force balance on the hand constraint enables us to compute the force control magnitude $\eta_f$, and the passive reaction forces $\eta_v$ in the velocity control directions. 

Our HFVC synthesis algorithm is similar to~\cite{hou2021efficient}, but we introduce several modifications that make the algorithm feasible for our domain.
First, we do not require pre-designed feasible trajectories for the HFVC to follow.
These trajectories are typically manually engineered or generated with simulations~\cite{cheng2021contact} with ground truth object feedback, which is impractical in our domain.
Instead, we interpolate a trajectory from start and goal object pose.
While this will result in infeasible object trajectories that would sometimes cause execution failures, our learned precondition function would classify this and help the planner avoid such executions.
Second, compared to the quasi-static assumption in~\cite{hou2021efficient}, we also allow short dynamic motions to allow object dropping. 
Nevertheless, the quasi-dynamic assumption still prevents us from doing fully dynamic manipulation.
However, our models are more limited than the general robot model in \cite{hou2021efficient}. 
The action space of the robot is just 6D end-effector task space, assuming the end-effector making one sticking point contact or a sticking patch contact. 
We are also using maximum velocity control while \cite{hou2021efficient} allows the users to choose from maximum or minimum velocity control. 

\subsection{HFVC execution}

\begin{table}[!h]
\centering
\begin{tabular}{l|l|l|l}
\hline
           & $S_v$   & $S_f$ & $D$                           \\ \hline
Simulation & $300I$  & $I$   & $diag([10, 10, 10, 1, 1, 1])$ \\
Real World & $1800I$ & $2I$  & $diag([10, 10, 10, 1, 1, 1])$
\end{tabular}
\caption{
    \footnotesize
    Gains used by low-level controller to follow HFVC targets in simulation and real world.
}
\label{tab:gains}
\end{table}

To execute a given HFVC command $h$ through our torque control scheme, our low-level controller first compute the current desired generalized force, velocity $f_e \in \mathbb{R}^6$ and $v_e \in \mathbb{R}^6$ from $h$: $f_e = T_f \eta_f$, and $v_e = T_v \eta_v$.
These are then used to compute commanded robot torques:
\begin{align}
    \tau = J^\top(S_vv_e + S_f f_e - Dv)
\end{align}
where $S_v$, $S_f$, and $D$ are diagonal matrices, the first two correspond to velocity and position error gains, and the third is for a damping term to achieve smoother control with $v$ being the generalized end-effector velocity.
See Table~\ref{tab:gains} for values of these matrices.
For simplicity, we did not write terms that correspond to compensating for gravity and Coriolis forces.

The HFVC skill terminates if both the position and rotational difference between the estimated object pose and the subgoal pose do not improve for more than $0.1$ seconds.

\subsection{HFVC Precondition Learning}

\begin{figure}[!h]
    \centering
    \includegraphics[width=\linewidth]{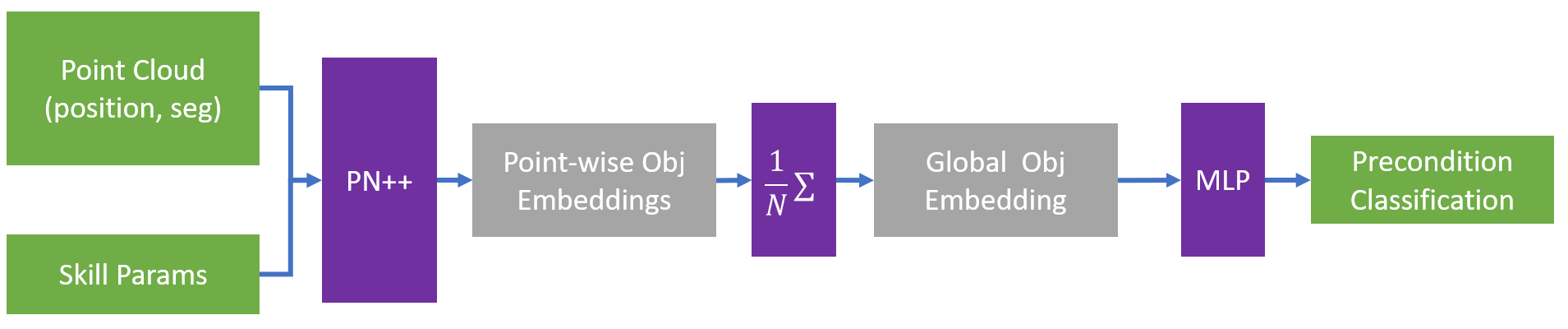}
    \caption{
        \footnotesize
        Precondition learning model architecture
    }
    \label{fig:model}
\end{figure}

For sampling object scales during data generation, we apply random non-uniform scaling of a range $[0.6, 1.3]$ along each principal axis, with objects that have scales near the $0.6$ and $1.3$ being used in the test set (~$10\%$ of the $440$k data collected ) are in the test set.

During training, we randomly subsample $256$ object points and $256$ environment points to pass into our network.
Because the object tends to be much smaller than the environment, this subsampling ensures the network sees a sufficient amount of object points relative to environment points.

Training the model takes about $20$ epochs for validation loss to converge, which takes about $100$ hours on an Nvidia A6000.

The following is a list of things we experimented with to improve precondition prediction performance that ultimately had negligible or even negative effects:
\begin{enumerate}
    \item \textbf{Adding point cloud normals as an input feature.} While this had no effect on performance on simulation data, it drastically worsened real-world precondition predictions, because real-world normals, even ones extracted from TSDFs, are very noisy.
    \item \textbf{Adding point-wise differences to environment points as an additional feature.} This is an $\mathbb{R}^3$ vector that starts from an object point and ends in the closest environment point such that the vector is also within $30^\circ$ of the point's normal direction. The motivation for this feature is to better capture object-environment contact relationships, but this had negligible effects on learning performance.
    \item \textbf{Using robust labels.} Instead of labeling a (point cloud, parameter) tuple as positive or negative by the result of running HFVC on that parameter, a robust label is generated by running many HFVC trials on perturbed parameters near the original parameter and labeling the original positive if more than $50\%$ of these noisy executions also satisfy preconditions. The motivation for this is to obtain cleaner and focused data that may be simpler to learn. However, using robust labels had negligible effect on performance, and generating the data for these labels took much longer than just using a single execution.
    \item \textbf{Adding parameter features to the object embeddings after PointNet++, instead of adding it to each object point before.} This change actually made performance slightly worse.
    \item \textbf{Using point clouds to represent parameters.} This means having 2 sets of point clouds, both parsed by PointNet++ and their outputs combined to make a prediction. The first set contains the environmental points, the initial object points, and the robot end-effector points placed at the initial robot-object contact point. The second set of points contain the environmental points, the object points where the object is placed at the subgoal, and corresponding end-effector points at the subgoal. The motivation for this representation is to allow the network to more directly parse geometric relationships. However, this change made performance slightly worse and because it needs to process more points, it made training much slower.
    \item \textbf{Adding contrastive losses.} We experimented with adding contrastive losses to the global object embedding to contrast between object types and parameter types (see Section~\ref{subsec:data} how these are defined). These losses had negligible effect on prediction performance.
    \item \textbf{Dynamics Learning.} We tried training the model to jointly predict preconditions and skill-level dynamics. The motivation for this is that 1) the richer training signal from dynamics losses may improve precondition learning and 2) the learned dynamics prediction may be more accurate than subgoals, leading to better planning performance. For 1), we found predicting dynamics has negligible effect on precondition performance. While our learned dynamics does indeed have much better aggregate next-state prediction errors (ADD $2.3$cm, see Figure~\ref{fig:model_dyn}) than directly using subgoals (ADD $3.4$cm, see Figure~\ref{fig:data_dyn}), it led to worse planning performance ($59.1\%$, compared to $73.2\%$ achieved by using only precondition model and subgoal dynamics). 
    
    We tried many variants of dynamics learning --- predicting 6D poses ($\operatorname{SE}(3)$), predicting 3D poses ($\operatorname{SE}(2)$) from the subgoal (the reached pose is in the same stable pose as the subgoal pose, otherwise the preconditions would not be satisfied), computing dynamics loss from pose predictions, and computing dynamics loss from the Average Discrepancy Distance (ADD, this is the mean distance between corresponding points of two point clouds). While some variants performed better than others, they were not able to result in better planning performance.
    
    Please see more detailed results in Section~\ref{subsec:data}, \ref{subsec:dyn}, and~\ref{subsec:plan}.
\end{enumerate}

\subsection{HFVC Precondition Sim-to-Real Gap}

To characterize the sim-to-real gap of the learned precondition function, We performed $61$ skill executions across the real world objects in Figure~\ref{fig:domain} and recorded the prediction results in Table~\ref{tab:training_results_rw}.
We show both the numbers for our approach and No-Params in simulation (from Table~\ref{tab:training_results}) and their performance on the real-world dataset (RW).
There is a $10\%$ accuracy drop for our approach and a much larger $21\%$ drop for No-Params.
Most of the accuracy drop came from reduced precision, meaning there are many samples where the model predicted were successful but in reality were not.
Recall actually increased for our approach, meaning a higher proportion of actual positives were predicted to be positive.
We note this is generally an acceptable trade-off for planning, as our replanning scheme allows recovering from false positive errors, but having high false negatives would prevent the planner from finding feasible solutions when they exist.
Lastly, we note that due to the degraded performance of the visual perception module on real-world data, the real-world precondition labels, which rely on real-world object pose estimates, are likely to be noisy, so we expect our models to actually perform better than these numbers would suggest.

\begin{table}[!t]
\centering
\begin{tabular}{l|l|l|l|l}
\hline
          & Ours   & Ours RW  & No-Params & No-Params RW \\ \hline
Accuracy  & $79\%$ & $69\%$   & $65\%$    & $44\%$          \\
Precision & $78\%$ & $53\%$   & $60\%$    & $35\%$             \\
Recall    & $77\%$ & $86\%$   & $79\%$    & $71\%$             \\
\end{tabular}
\caption{
    \footnotesize
    Precondition prediction results with different input representations on real world (RW) data.
}
\vspace{-10pt}
\label{tab:training_results_rw}
\end{table}

\section{Planner Details}
We use a planning algorithm similar to Real-time A* but with a few modifications that make the planner greedier but faster.
The planner has a planning budget of $30$ seconds and expands at most $10$ nodes with a maximum search depth of $3$.
While we only expand at most $10$ nodes, this actually requires generating many more successors because of the high branching factor (can be more than $100$).
To improve planning speed, we prune parameters that reach the same subgoal pose by only keeping one of them (there are many parameters that reach the same subgoal pose via different robot-object contact points).
This choice is made by sampling a probability distribution that is proportional to their predicted precondition satisfaction probabilities, where weights are computed via softmax.
In addition, we also terminate the search after a path to goal is found, instead of waiting until the goal state is removed from the open list.
This is because due to inaccurate transition models, we are not confident in the actual cost improvements of other plans, and we prefer to executing the first action then replan as needed.
While our algorithm is coded as a graph search problem, in practice it more closely resembles a tree search, because only a very rare occasions do two generated successor states coincide to be the same state (have the same object pose).

The goal pose is satisfied if the object reaches within $1.5$cm and $10^\circ$ tolerance.
The cost of each planning edge is $c = d + 0.5\theta$, where $d$ is the distance between the initial and reached object poses, and $\theta$ the angle.
To guide the search, we use the heuristic $pD(\prescript{W}{}{\Tau}_O, \prescript{W}{}{\Tau}_{g}) + (1-p) 1$, where $p$ is the prediction precondition satisfaction probability of the action that corresponds to the edge before the current state, $\prescript{W}{}{\Tau}_O$ is the object pose of the current planning state, and $\prescript{W}{}{\Tau}_g$ the goal pose.
Here, $D$ computes the distance between two poses the same way the cost is computed above.
The $(1-p) 1$ term can be thought of as applying a penalty of $1$ if the skill fails.
This heuristic is inadmissible, but we found it works well in practice when coupled with replanning.

\section{Experiment Details}

\subsection{Task Domain Setup}
The Franka Arm gripper is outfitted with finger extensions --- this helps the robot to reach farther into the shelf without incurring collisions.
The finger extensions are covered with Plasti Dip to increase friction.
From both simulated and real-world robot arms we receive feedback on end-effector poses, velocities, and external force-torques.
To obtain point clouds of the scene, in simulation we directly fuse together deprojected depth images from 3 simulated cameras whose combined viewpoints cover the entire shelf area.
In the real-world, this is done by capturing a series of depth images from a wrist-mounted RealSense D415 camera, and a final point cloud is produced via TSDF Fusion with Open3D.
See Figure~\ref{fig:domain} for an illustration.

\textbf{Pick-and-Place Skill.}
In order to more efficiently complete the task, we also use a Pick-and-Place skill that can grasp and move objects to subgoal poses.
The parameter for the Pick-and-Place skill contains two elements --- an initial antipodal grasp pose for the robot and a final placement pose for the object.
Grasp poses are generated from the estimated object primitives, and placement poses are sampled in a grid on the shelf surface.
We additionally sample neighboring stable poses and planar rotations for object placement poses.
These parameters are filtered by collision checking --- the robot should not collide with the environment and the grasp and placement poses, and the object should not collide with the environment at the placement pose.

\subsection{Data Generation Results}
\label{subsec:data}

\begin{figure}[!t]
    \centering
    \includegraphics[width=0.49\linewidth]{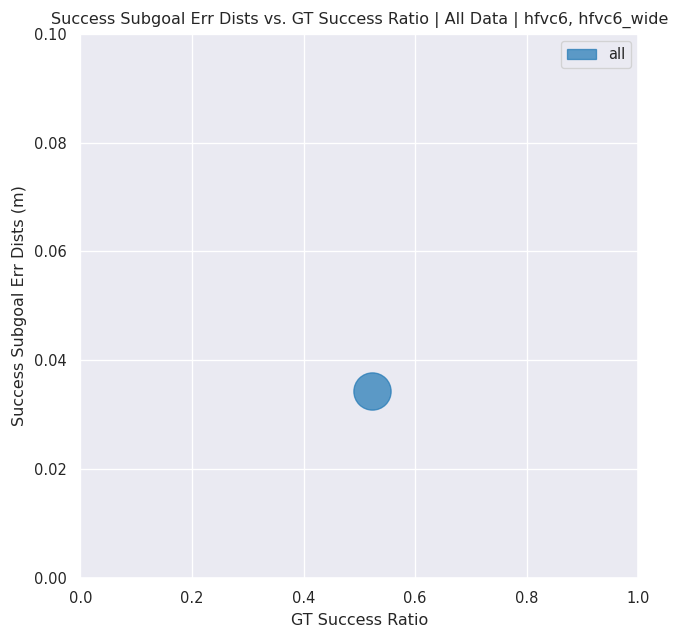}
    \includegraphics[width=0.49\linewidth]{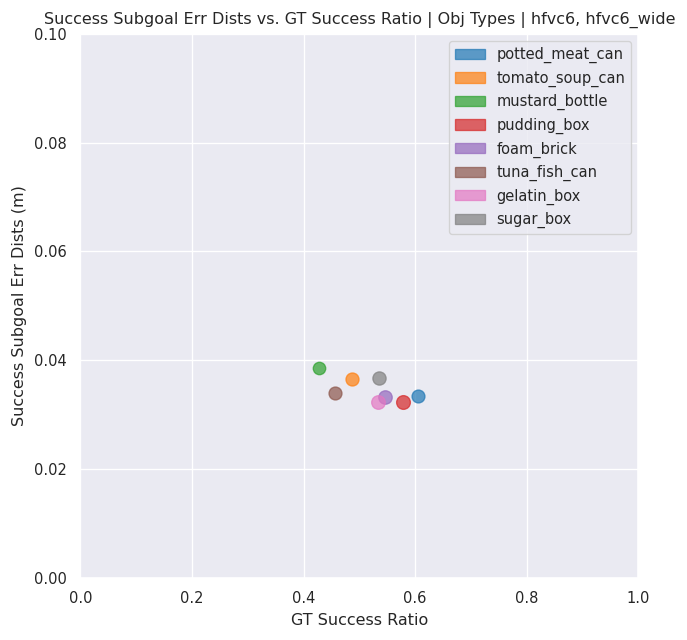}
    \includegraphics[width=0.49\linewidth]{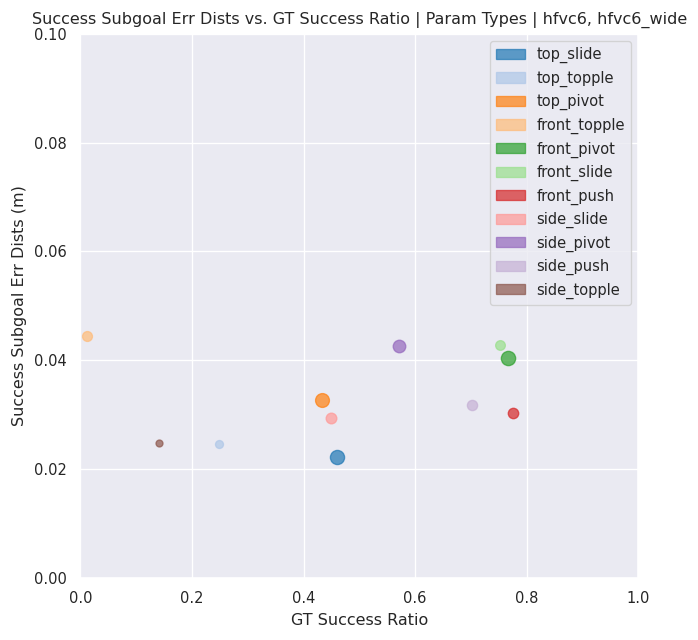}
    \includegraphics[width=0.49\linewidth]{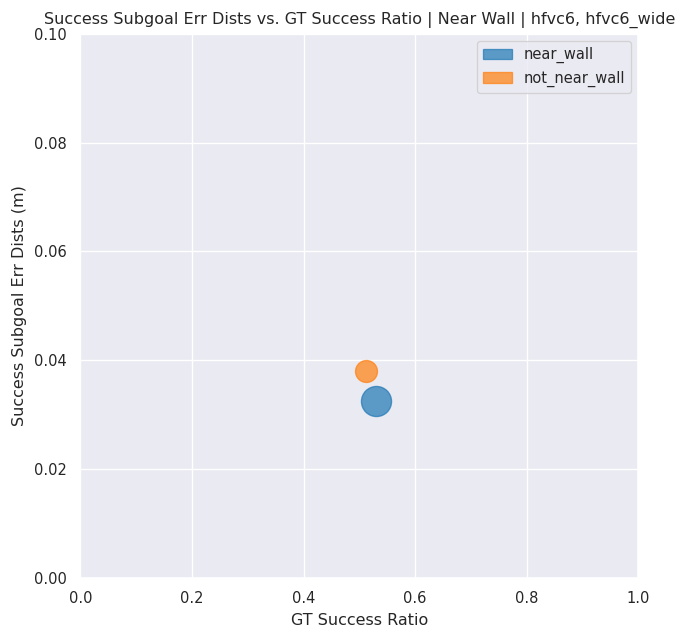}
    \caption{
        \footnotesize
        Precondition data generation statistics across different data segments.
        X-axis is the precondition success ratio.
        Y-axis is the ADD between the subgoal pose and the actual reached pose after skill execution; this is only aggregated over executions that satisfy preconditions.
        The area of each circle corresponds to the ratio of data size --- the number of data points in for that type of data divided by the the number of all data points.
        The four plots shows the these statistics across different data segments.
        Top left: all data.
        Top right: separated by object type.
        Bottom left: separated by parameter type.
        Bottom right: separated by whether or not the initial pose was near a shelf wall.
    }
    \label{fig:data_pc}
\end{figure}

\begin{figure}[!t]
    \centering
    \includegraphics[width=0.49\linewidth]{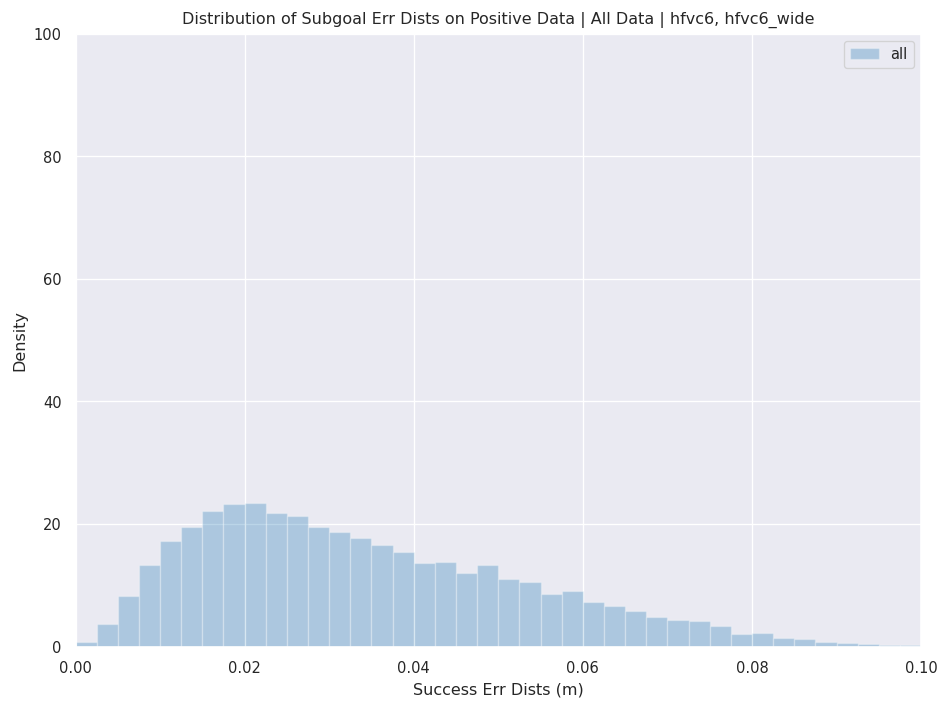}
    \includegraphics[width=0.49\linewidth]{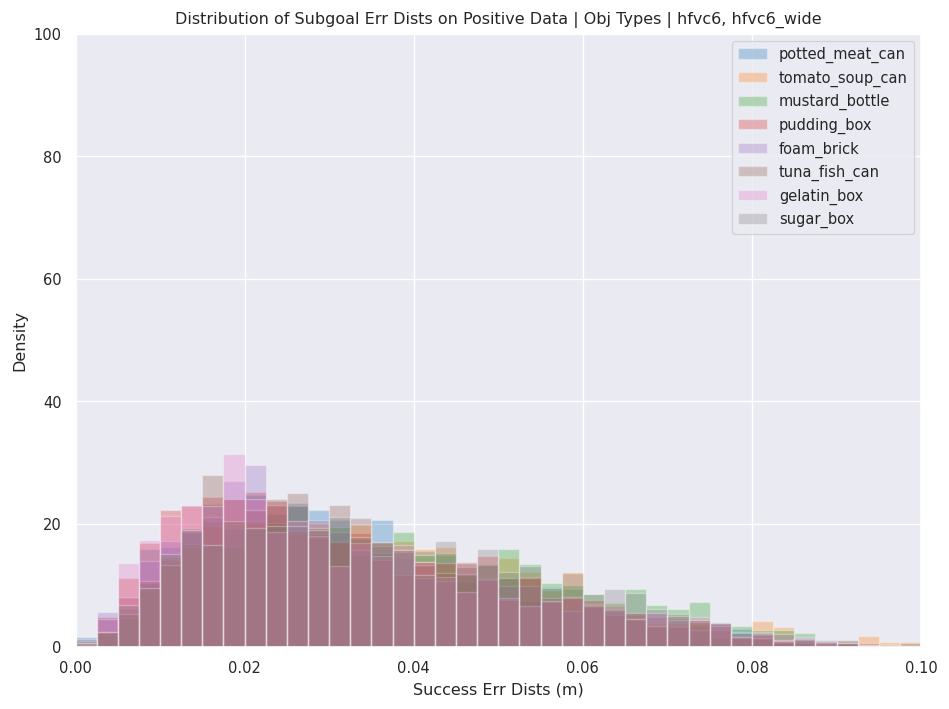}
    \includegraphics[width=0.49\linewidth]{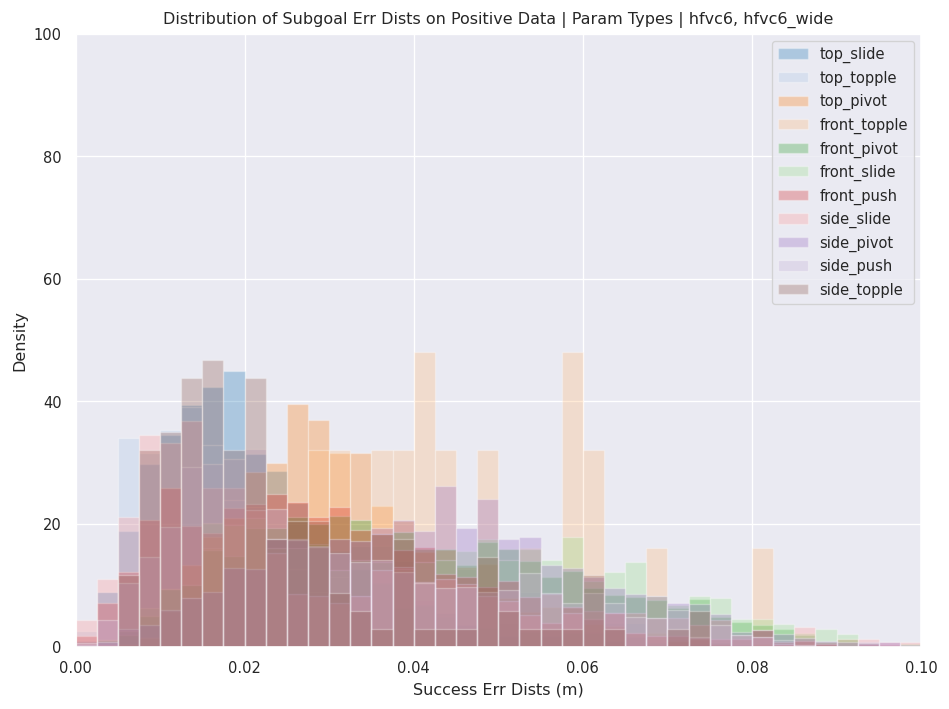}
    \includegraphics[width=0.49\linewidth]{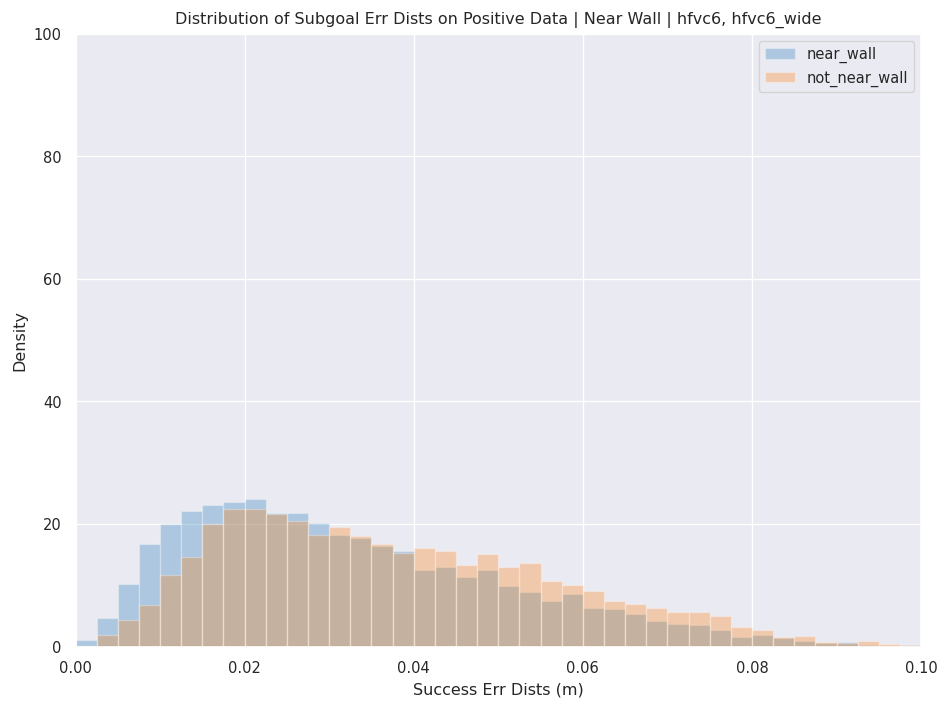}
    \caption{
        \footnotesize
        Subgoal dynamics error distributions across different data segments.
        We plot histograms of the ADD between subgoal and actual reached pose after skill execution; this is only aggregated over executions that satisfy preconditions.
        The four plots shows the these statistics across different data segments.
        Top left: all data.
        Top right: separated by object type.
        Bottom left: separated by parameter type.
        Bottom right: separated by whether or not the initial pose was near a shelf wall.
    }
    \label{fig:data_dyn}
\end{figure}

In Figures~\ref{fig:data_pc} we plot precondition success ratios and the mean subgoal pose error ADDs; in Figure~\ref{fig:data_dyn} we plot the distribution of subgoal pose error ADDs.
Pose errors are only computed over executions that satisfy preconditions.
In both figures, we separate the aggregate statistics into different data segments.
One is across all objects, one is across the $8$ YCB objects we used, one is across parameter types, and one is across whether or not the initial state is close to a shelf wall.
Parameter type is denoted by concatenating an adjective (front, top, side) and a verb (slide, push, pivot, topple).
The adjective refers to the position of the initial robot-object contact point, whether it is in front of the object (toward the shelf opening), top of the object, or the side of the object.
The verb refers to the motion achieved by the combination of the robot-object contact point and the desired subgoal pose.
For slide and push, the subgoal pose only differs from the initial pose via translation, while pivot and topple have rotation components.
The difference between slide and push is that comparing to the delta pose direction, the robot-object contact normal is more orthogonal for slide than it is for push.
The difference between pivot and topple is that pivot does not change the stable pose, while topple does.
We compute these parameter types only for debugging and visualization purposes --- they are not used by the precondition model.

\subsection{Precondition Learning Results}
\label{subsec:pc}

\begin{figure}[!t]
    \centering
    \includegraphics[width=\linewidth]{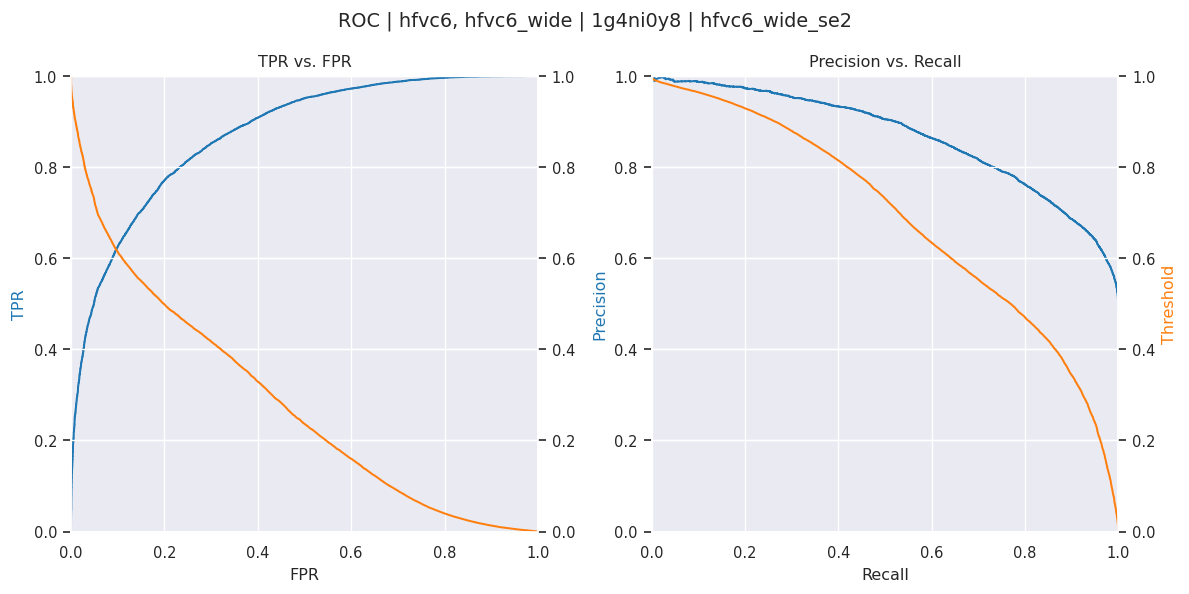}
    \caption{
        \footnotesize
        Precondition performance curves.
        Left: True Positive Rate vs. True Negative Rate.
        Right: Precision vs. Recall.
        Points on the threshold curves correspond to the threshold used for the corresponding point (same x-axis location) on the blue curves, where the threshold is the y-axis value on the right.
    }
    \label{fig:model_pc_perf}
\end{figure}

\begin{figure}[!t]
    \centering
    \includegraphics[width=0.49\linewidth]{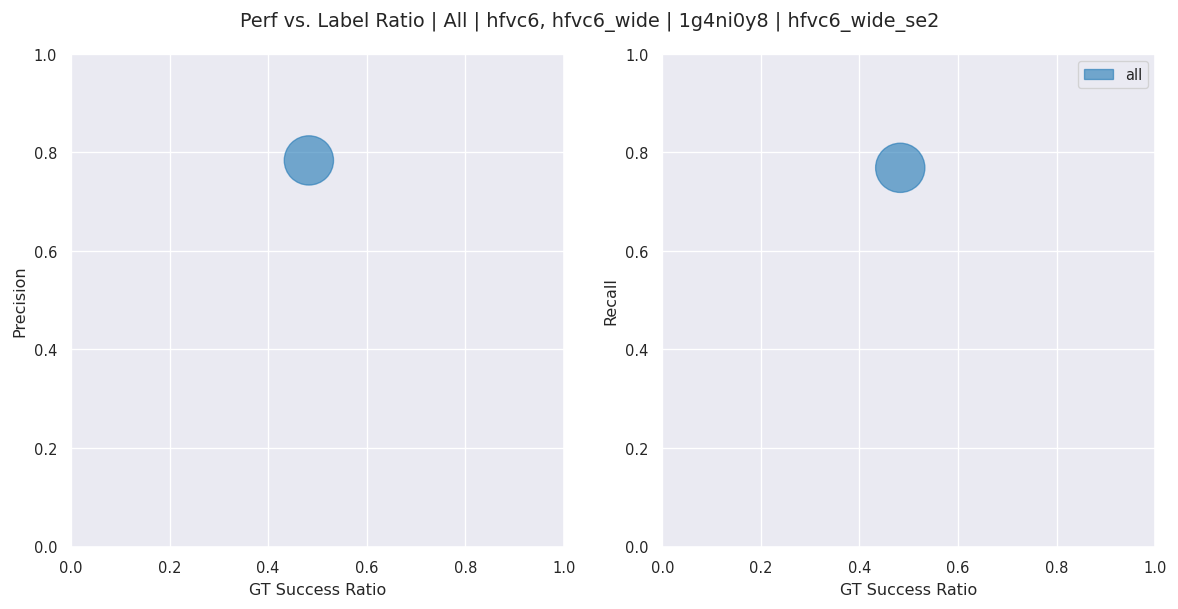}
    \includegraphics[width=0.49\linewidth]{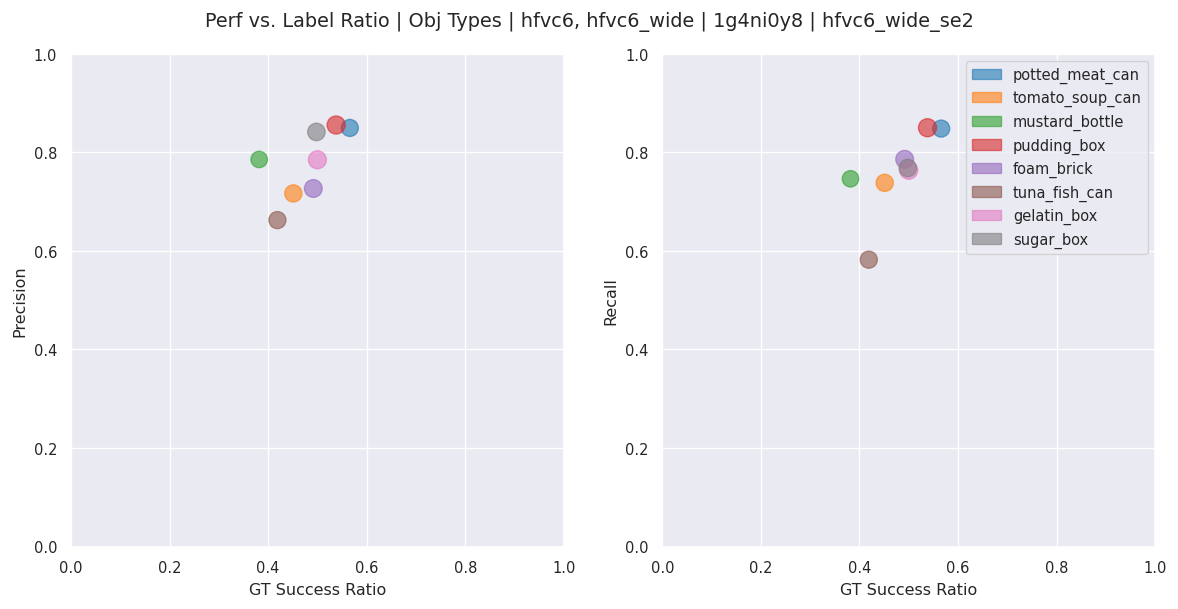}
    \includegraphics[width=0.49\linewidth]{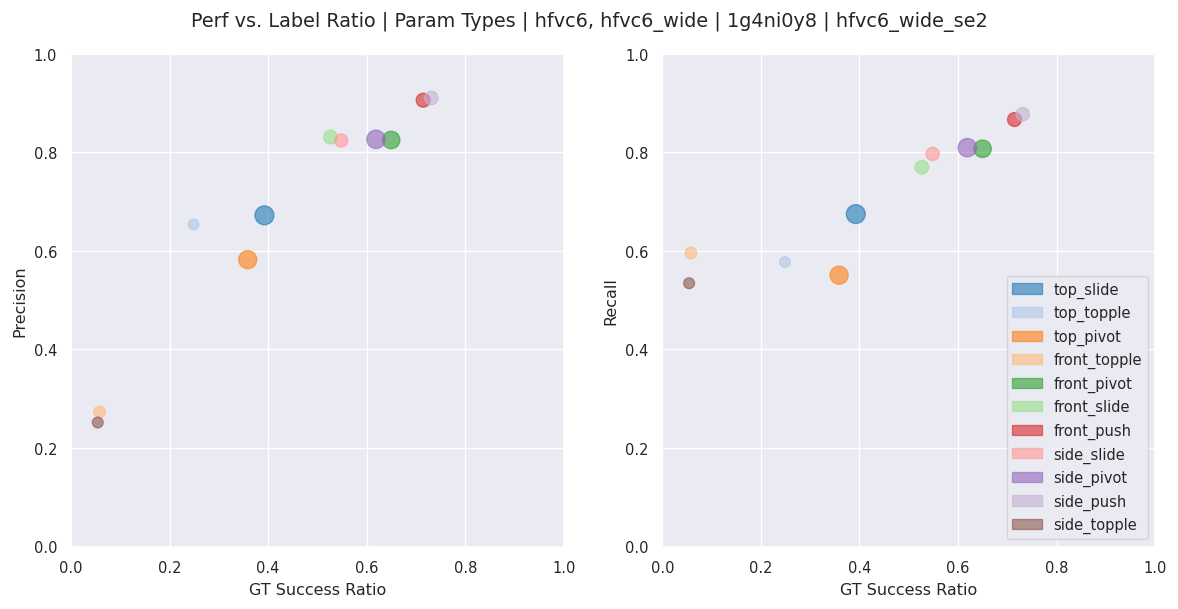}
    \includegraphics[width=0.49\linewidth]{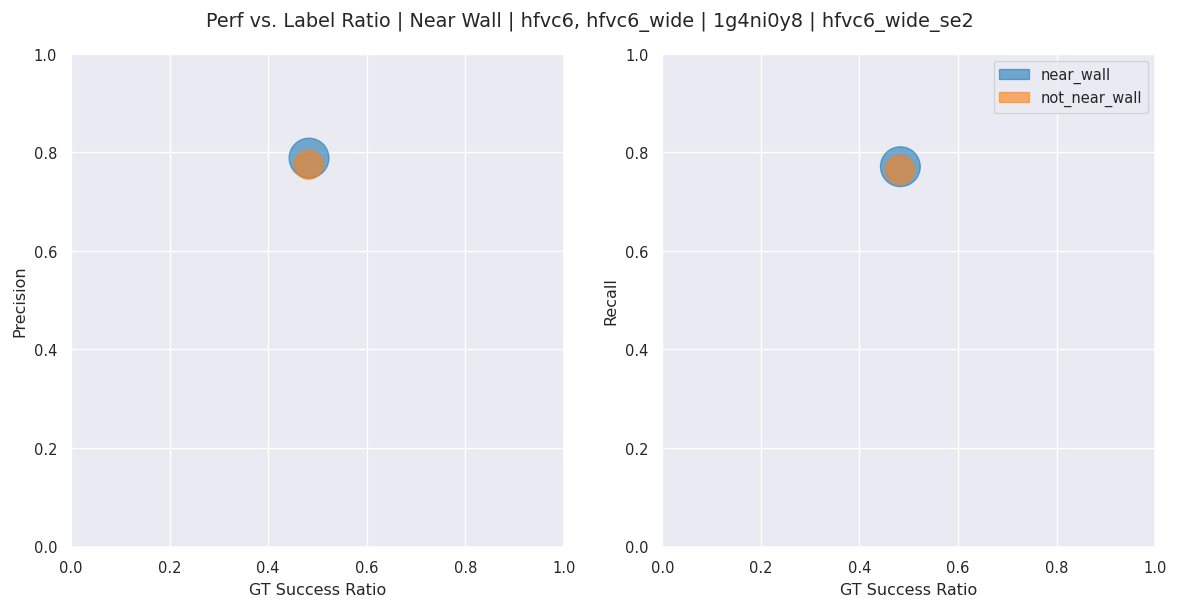}
    \caption{
        \footnotesize
        Precondition precision, recall vs. ground truth precondition success ratio across data types.
        X-axis is the ground truth precondition success ratio.
        Y-axis plots precision and recall.
        The area of each circle corresponds to the ratio of data size --- the number of data points in for that type of data divided by the the number of all data points.
        The four plots shows the these statistics across different data segments.
        Top left: all data.
        Top right: separated by object type.
        Bottom left: separated by parameter type.
        Bottom right: separated by whether or not the initial pose was near a shelf wall.
    }
    \label{fig:model_pc}
\end{figure}

In Figure~\ref{fig:model_pc_perf} we plot precondition performance curves (True Positive Rate vs. True Negative Rate, Precision vs. Recall), and in Figure~\ref{fig:model_pc} we further breakdown Precision and Recall numbers into different data segments, similar to the data generation results above.

\subsection{Dynamics Learning Results}
\label{subsec:dyn}
\begin{figure}[!t]
    \centering
    \includegraphics[width=0.49\linewidth]{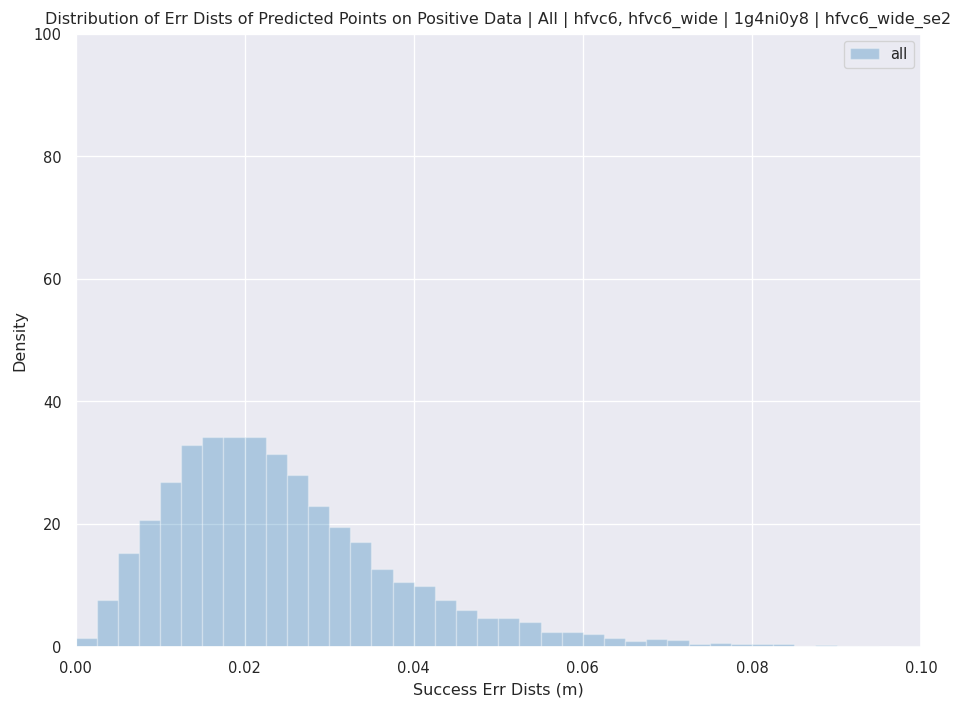}
    \includegraphics[width=0.49\linewidth]{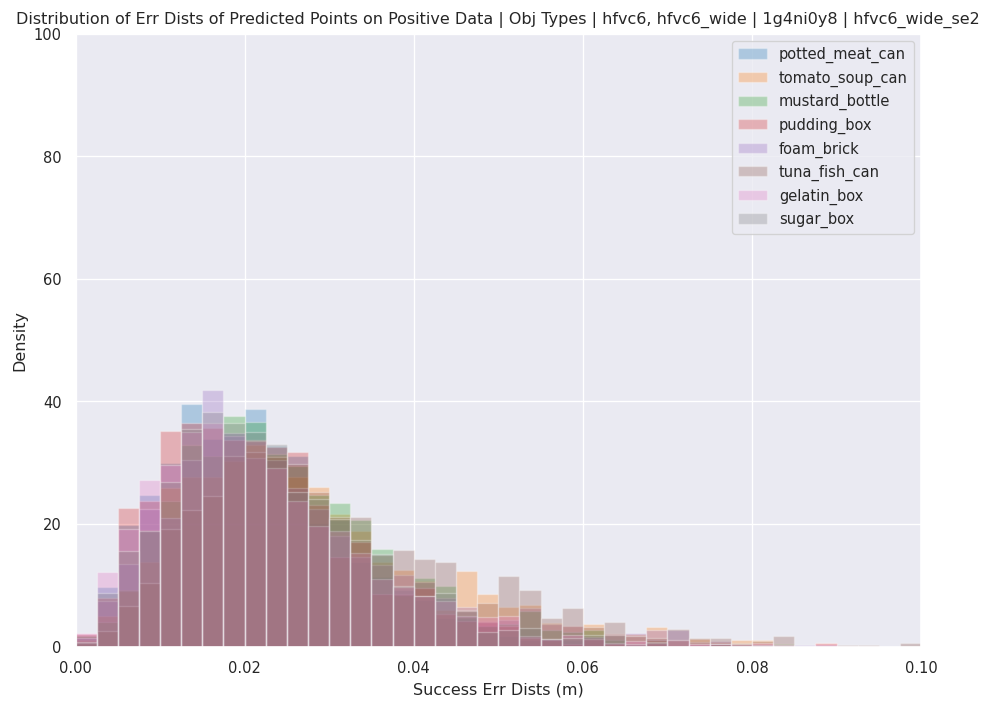}
    \includegraphics[width=0.49\linewidth]{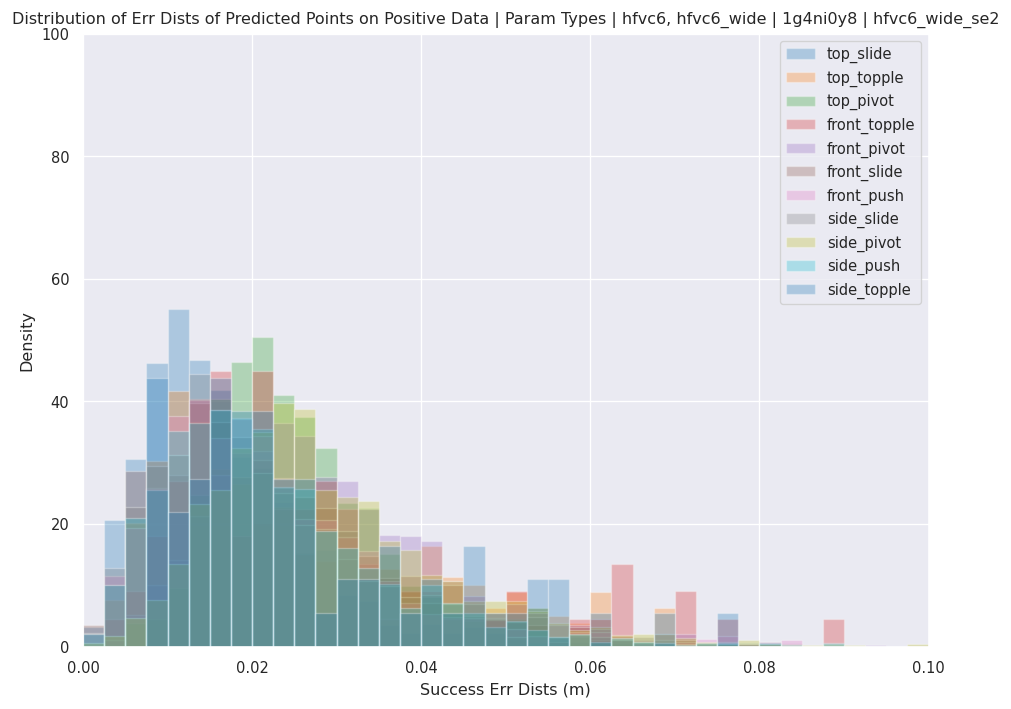}
    \includegraphics[width=0.49\linewidth]{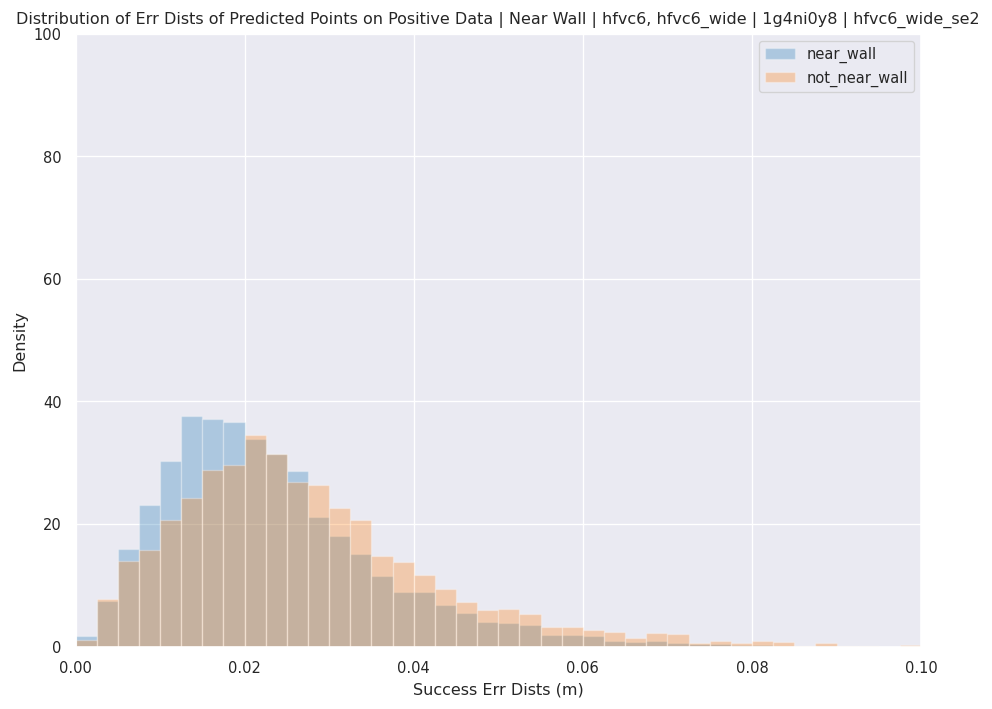}
    \caption{
        \footnotesize
        Predicted dynamics error distributions across different data segments.
        We plot histograms of the ADD between subgoal and actual reached pose after skill execution; this is only aggregated over executions that satisfy preconditions.
        The four plots shows the these statistics across different data segments.
        Top left: all data.
        Top right: separated by object type.
        Bottom left: separated by parameter type.
        Bottom right: separated by whether or not the initial pose was near a shelf wall.
    }
    \label{fig:model_dyn}
\end{figure}

For dynamics learning, we train a model to jointly predict next object pose and the preconditions by adding a new output head from the global object embedding.
To represent the next pose prediction, our model predicts a planar 3D delta pose (x-y translation and z-axis rotation) that when applied to the subgoal pose, should give the actual reached pose.
In experiments this yielded better prediction performance than directly predicting the 6D delta pose from the initial pose.
The dynamics loss is represented by the ADD between the object points at the predicted pose than at the actual pose.
This gave better prediction performance than directly regressing to the ground truth delta pose values.
Dynamics loss is only back-propagated for data points that have positive precondition labels.
See dynamics prediction results in Figure~\ref{fig:model_dyn}, where we plot distributions of predicted object pose errors over executions that satisfy ground truth preconditions.
Note that this distribution has lower mean errors and less outliers than directly using subgoals (Figure~\ref{fig:data_dyn}).

\subsection{Planning Results}
\label{subsec:plan}

\begin{figure}[!t]
    \centering
    \includegraphics[width=0.49\linewidth]{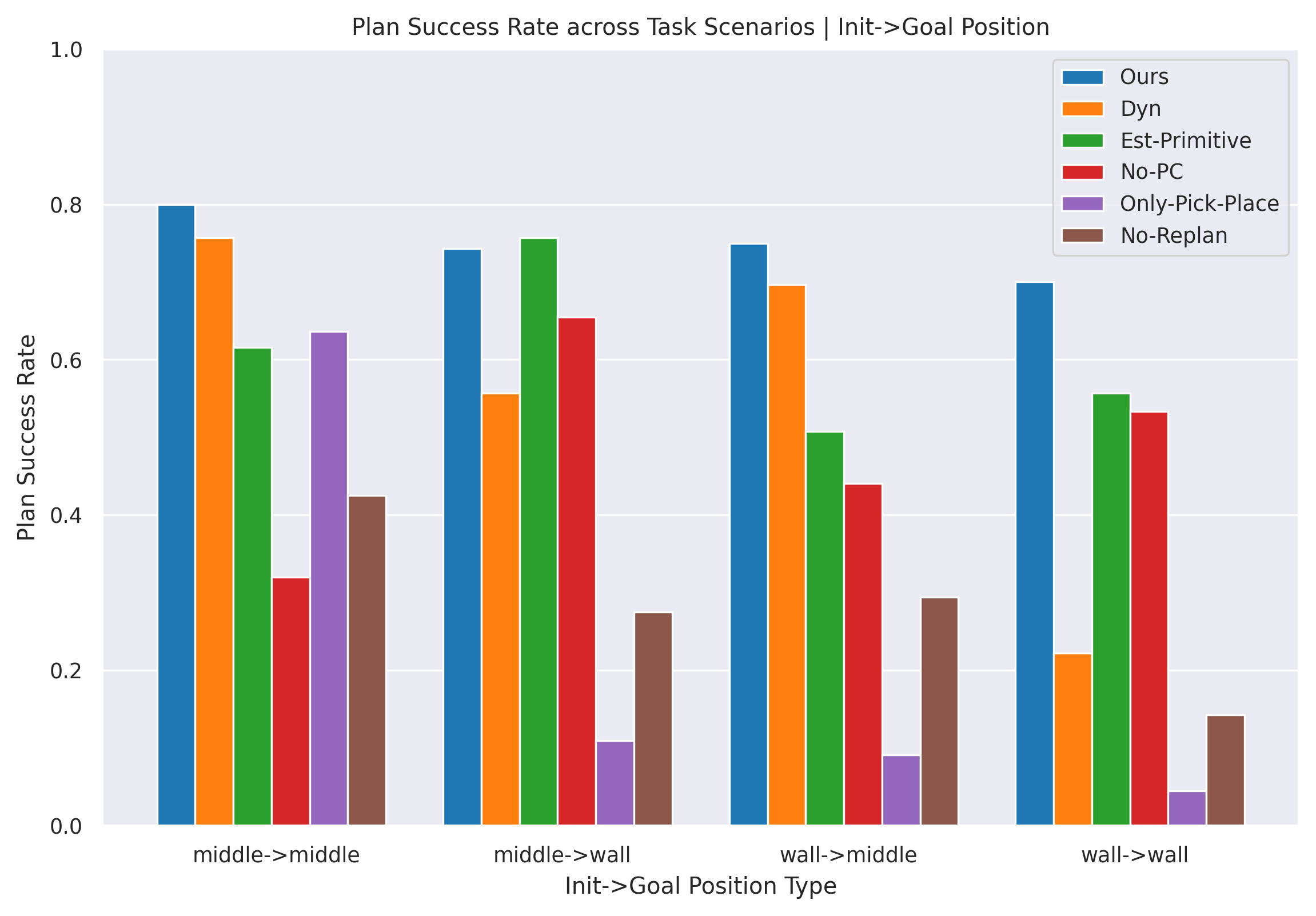}
    \includegraphics[width=0.49\linewidth]{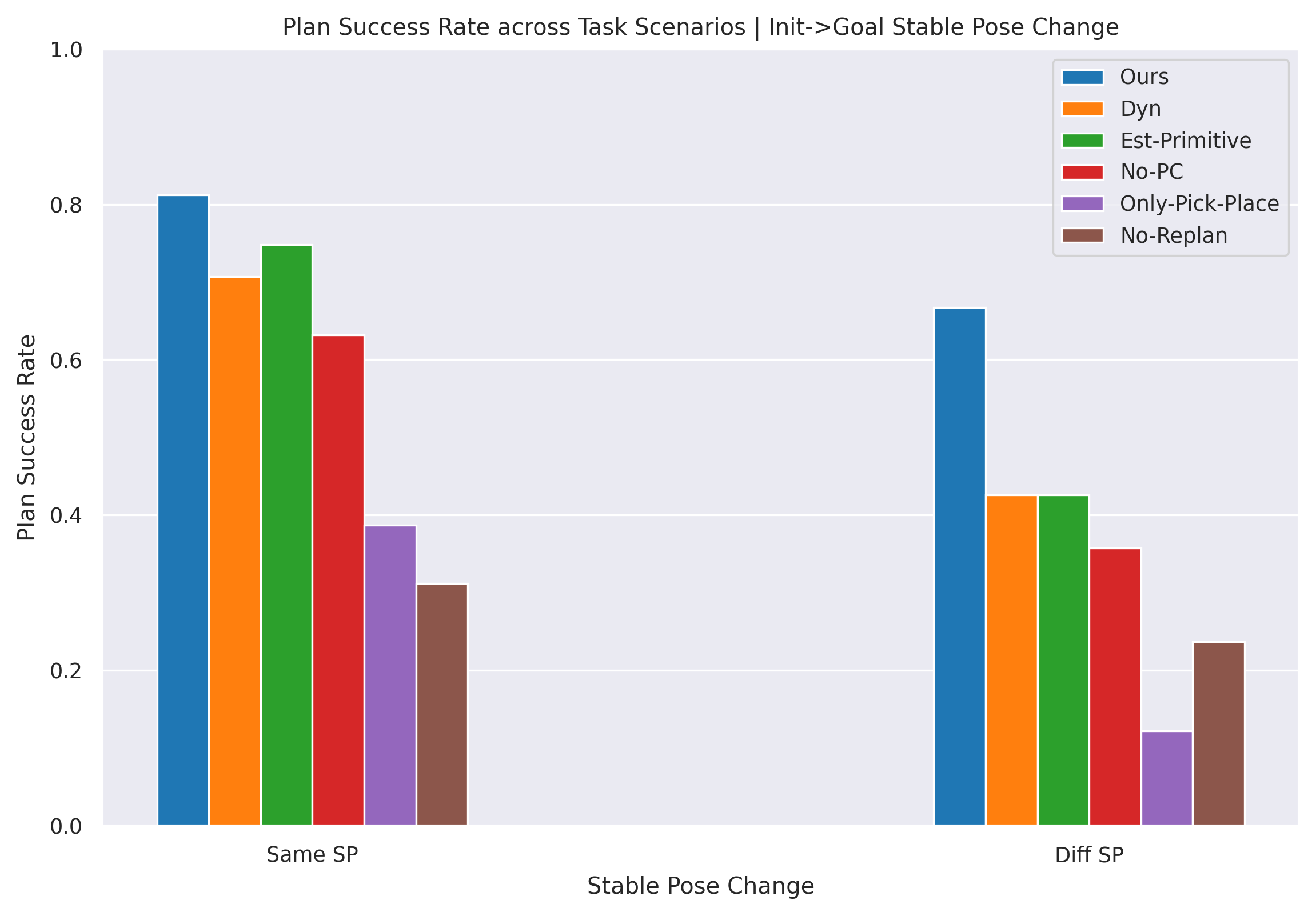}
    \caption{
        \footnotesize
        Task success rate for ablations across different task scenarios.
        Left: initial to goal object position types.
        Right: whether or not initial and object poses are in stable pose configurations
    }
    \label{fig:plan_success_by_task_scenarios}
\end{figure}

\textbf{Planning with learned dynamics.}
We report planning results of using a model that jointly predicts skill preconditions and dynamics (Dyn) in Table~\ref{tab:dyn}.
This achieves a success rate of $59.1\%$, which is lower than our method that uses only the learned preconditions with subgoals ($73.2\%$), but it is comparable to Est-Primitive ($61.1\%$).
Note that this method has a significantly longer total plan time (mean $89.1$s, as opposed to the mean of $43.1$s for our method), which is indicative of needing to replan more frequently.
In Figure~\ref{fig:plan_success_by_task_scenarios} (explained in more detail below), the most dramatic difference between our approach and Dyn is for tasks where both the initial and goal object poses are near shelf walls.
This points to the model's inability to sufficiently take into account object-environment interactions to accurately predict skill dynamics in these scenarios.

\begin{table}[!t]
\centering
\begin{tabular}{l|l|l|l}
\hline
    & Plan Success & Plan Time (s) & Plan Length \\ \hline
Dyn & $59.1\%$     & $89.1\pm60.9$ & $2.6\pm1.4$
\end{tabular}
\caption{
    \footnotesize
    Planning performance of using skill model that jointly predicts skill preconditions and dynamics.
}
\label{tab:dyn}
\end{table}

\textbf{Planning performance across task scenarios.}
To better understand the differences among the $4$ variants we compared in our main planning result, we plot success rates for all methods across different task scenarios in Figure~\ref{fig:plan_success_by_task_scenarios}.
We report two types task scenario comparisons, one on whether the initial and goal object poses are in the middle of the shelf or near the shelf walls, and another one on whether these two poses are in the same stable pose configurations.
For init$\rightarrow$goal position types, No-PC has the worst performance in middle$\rightarrow$middle.
This is because many HFVC motions, like pushing and sliding, are infeasible or not robust when the object is not close environmental constraints, so the planner is overly optimistic on the types of plans it can achieve.
Only-Pick-Place is able to find many successful plans for middle$\rightarrow$middle, but less so for other scenarios. 
This is expected as most object poses near shelf walls are not graspable.
No-Replan has much worse performance than our approach across the board, indicating that subgoal poses are not accurate enough to use as a transition model for multi-step planning.
While all methods perform worse when the stable poses change, our method is still able to complete over $60\%$ of the tasks, while the best alternative (Est-Primitive) is in the low $40\%$s.

\subsection{Example Plans}

\begin{figure}[!t]
    \centering
    \includegraphics[width=\linewidth]{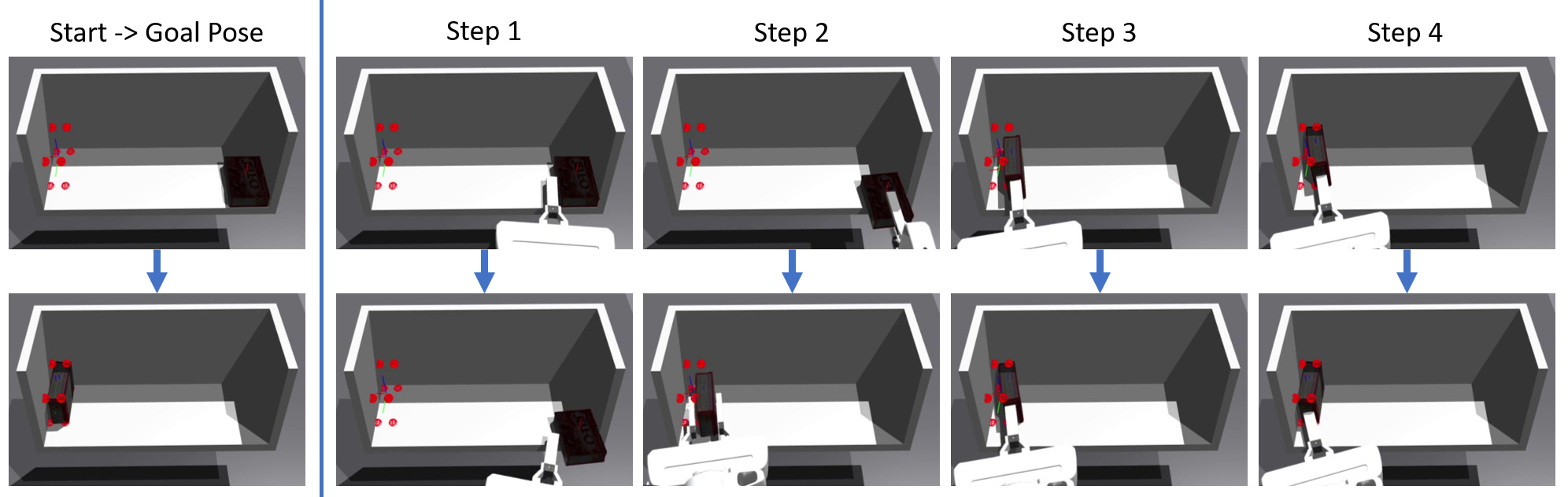}
    \caption{
        \footnotesize
        Example plan execution in simulation for the gelatin box.
        Left column is the start and goal state (note they are in different stable poses).
        The red circles indicate the intended goal pose, while the object on the bottom pictures is the reached final pose.
        Right columns are the series of HFVC and Pick-and-Place skills the planner found to complete the task.
        Note the use of HFVC skills in the beginning and in the end when the object grasps are occluded by shelf walls.
    }
    \label{fig:plans_sim_0}
\end{figure}

\begin{figure}[!t]
    \centering
    \includegraphics[width=0.8\linewidth]{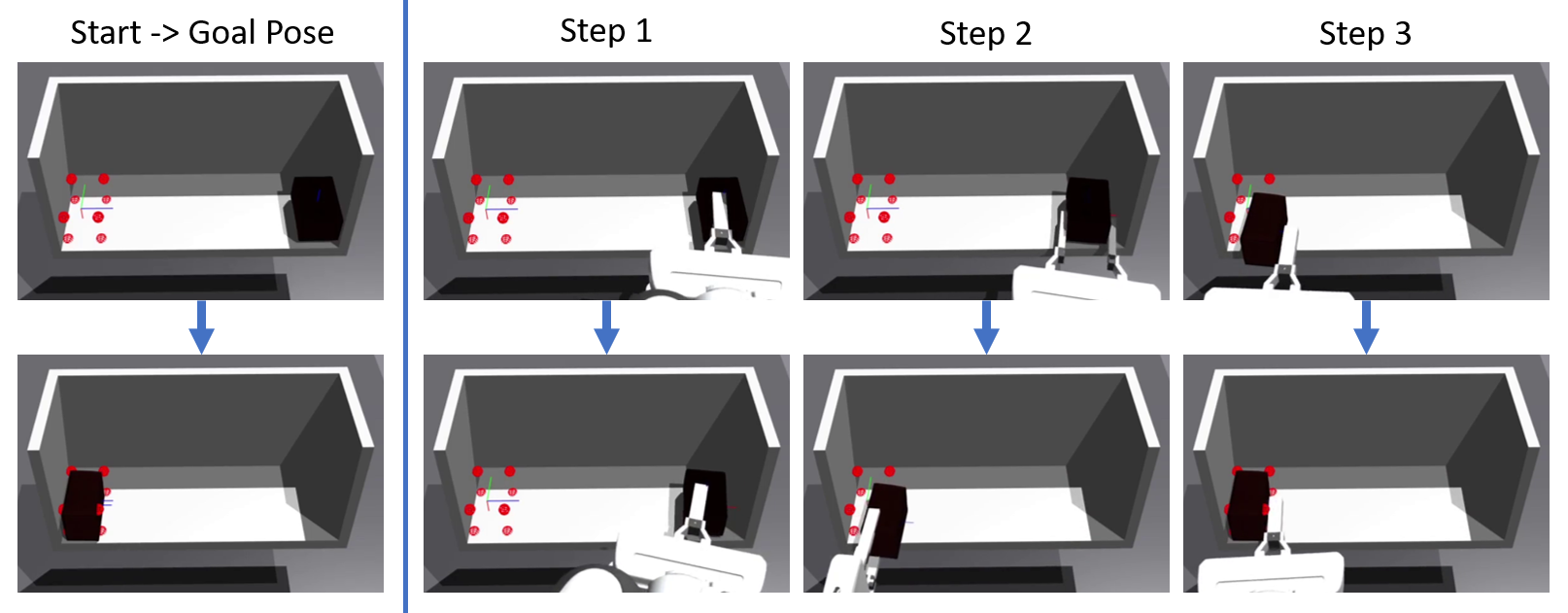}
    \caption{
        \footnotesize
        Example plan execution in simulation for the foam brick.
        Left column contains the start and goal states (note they are in different stable poses).
        The red circles indicate the intended goal pose, while the object on the bottom pictures is the reached final pose.
        Right columns are the series of HFVC and Pick-and-Place skills the planner found to complete the task.
        Note the use of HFVC skills in the beginning and in the end when the object grasps are occluded by shelf walls.
    }
    \label{fig:plans_sim_1}
\end{figure}

\begin{figure}[!t]
    \centering
    \includegraphics[width=0.8\linewidth]{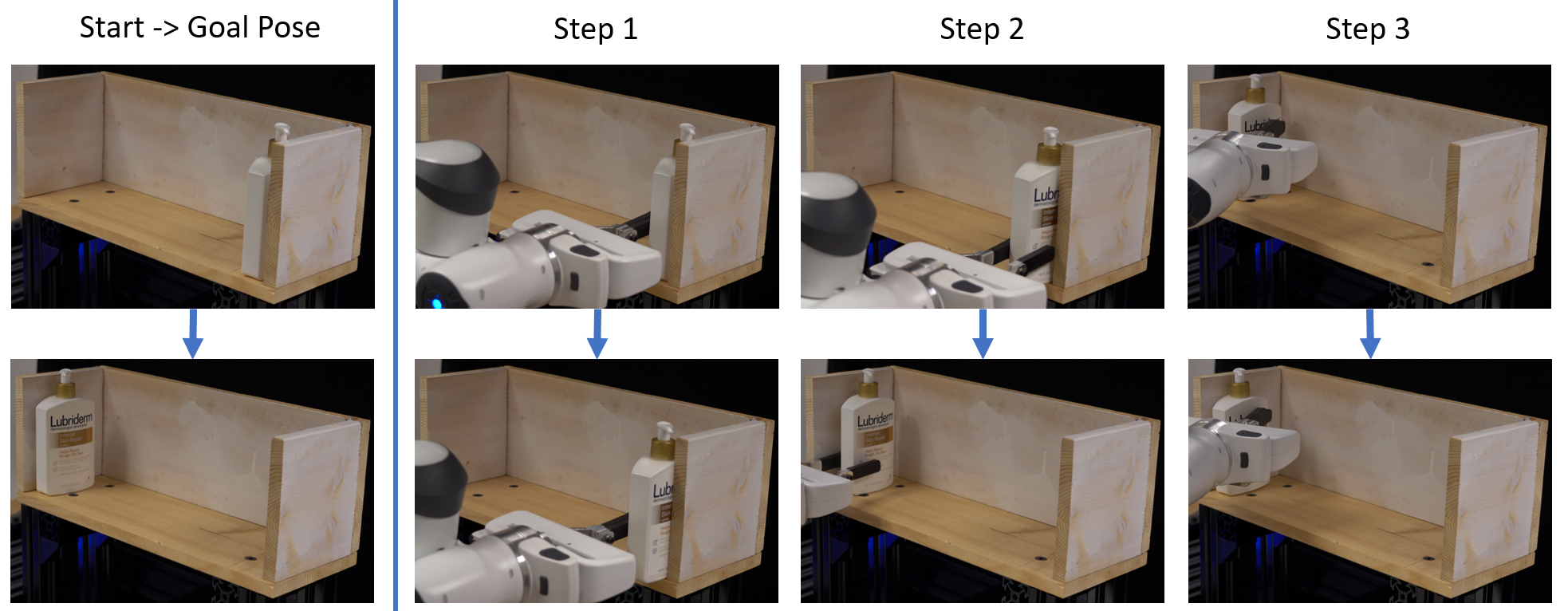}
    \caption{
        \footnotesize
        Example plan execution in the real world.
        Left column contains the start and goal states (note they are in different stable poses).
        Right columns are the series of HFVC and Pick-and-Place skills the planner found to complete the task.
    }
    \label{fig:plans_real_0}
\end{figure}

\begin{figure}[!t]
    \centering
    \includegraphics[width=0.8\linewidth]{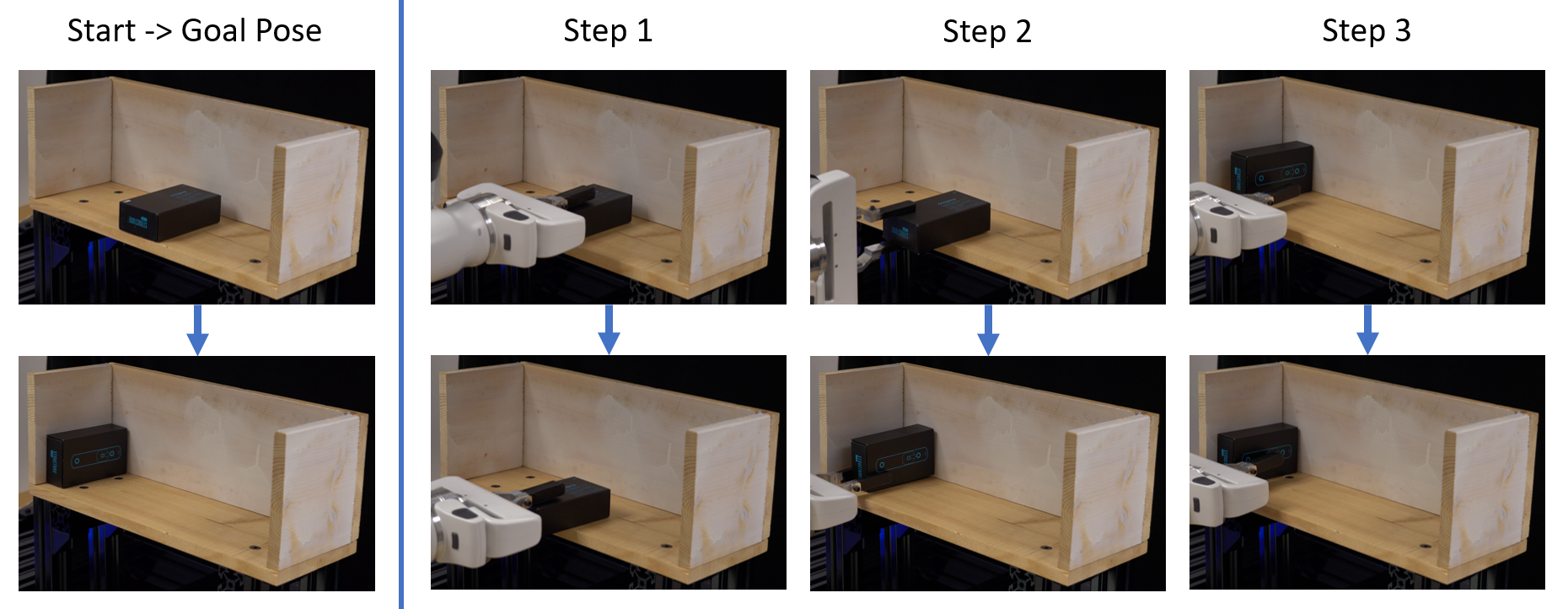}
    \caption{
        \footnotesize
        Example plan execution in the real world.
        Left column contains the start and goal states (note they are in different stable poses).
        Right columns are the series of HFVC and Pick-and-Place skills the planner found to complete the task.
    }
    \label{fig:plans_real_1}
\end{figure}

\begin{figure}[!t]
    \centering
    \includegraphics[width=0.75\linewidth]{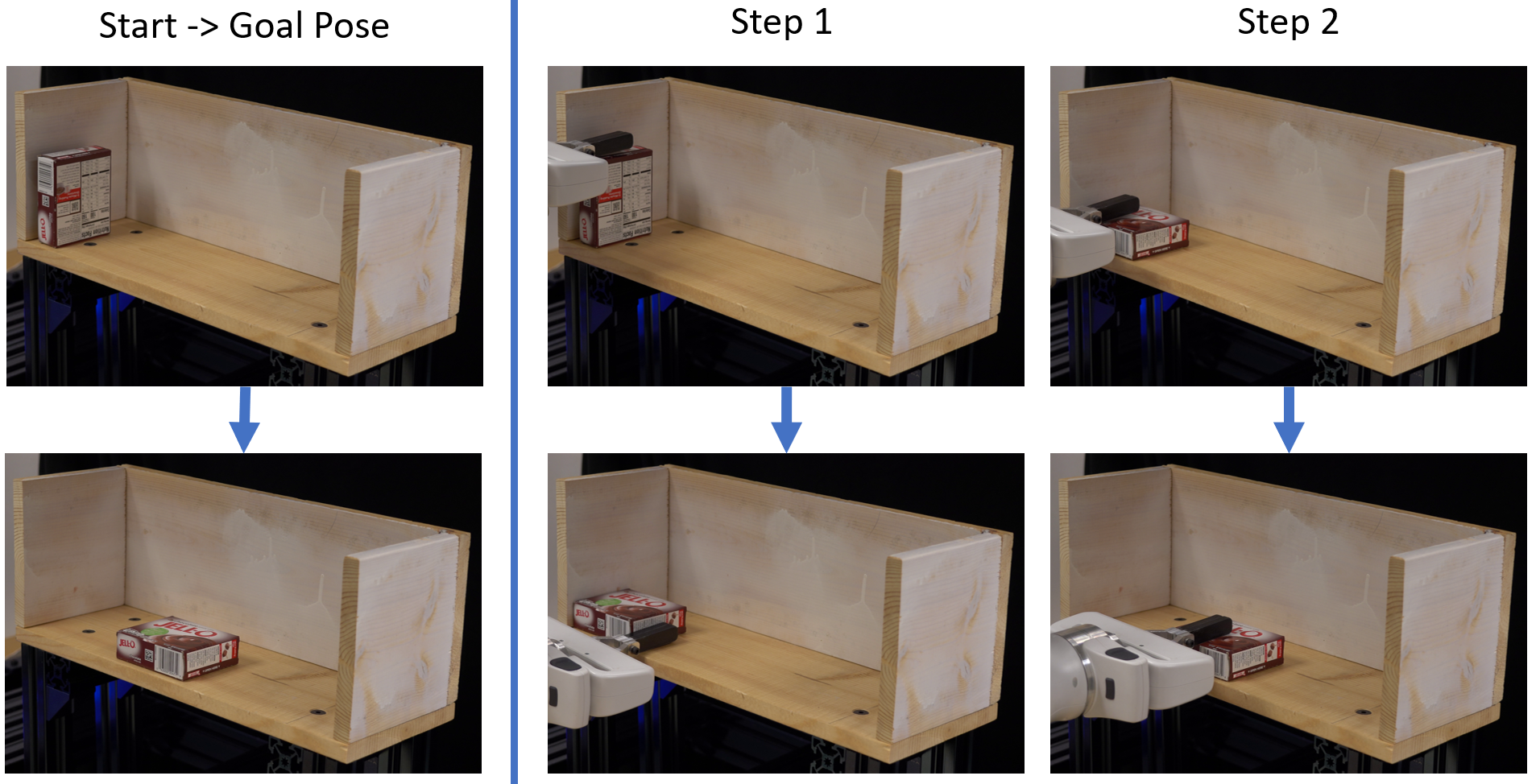}
    \caption{
        \footnotesize
        Example plan execution in the real world.
        Left column contains the start and goal states (note they are in different stable poses).
        Right columns are the series of HFVC and Pick-and-Place skills the planner found to complete the task.
    }
    \label{fig:plans_real_2}
\end{figure}

\begin{figure}[!t]
    \centering
    \includegraphics[width=0.75\linewidth]{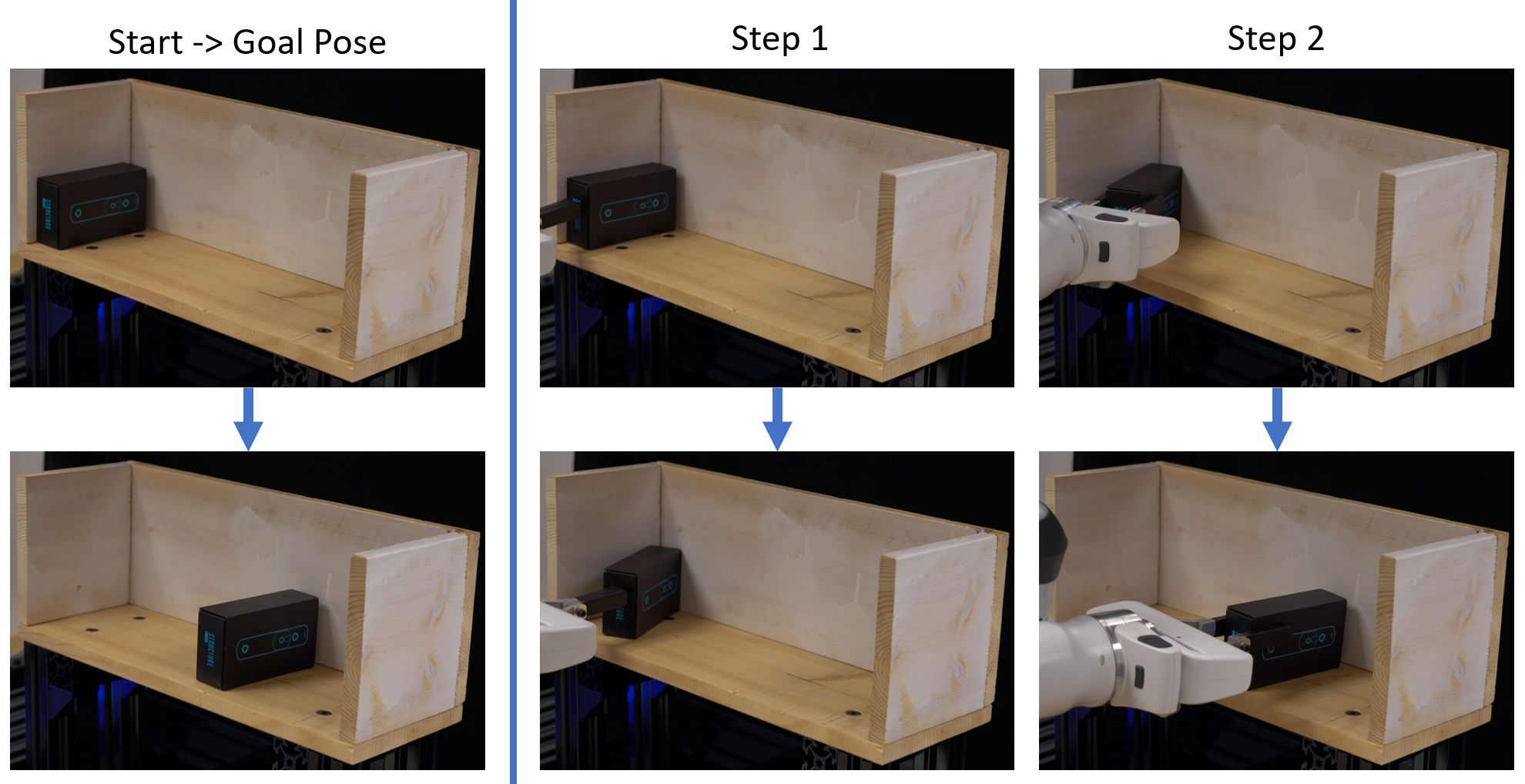}
    \caption{
        \footnotesize
        Example plan execution in the real world.
        Left column contains the start and goal states (note they are in different stable poses).
        Right columns are the series of HFVC and Pick-and-Place skills the planner found to complete the task.
    }
    \label{fig:plans_real_3}
\end{figure}

Figures~\ref{fig:plans_sim_0} and~\ref{fig:plans_sim_1} show two example plan executions in simulation.
In both cases, the object starts and ends near the shelf walls, so the robot needs to use HFVC skills to manipulate the object absence of available grasps.
Figures~\ref{fig:plans_real_0}, \ref{fig:plans_real_1}, \ref{fig:plans_real_2}, and~\ref{fig:plans_real_3} have example plan executions in the real world.
For videos and additional plans, please see the website~\url{https://sites.google.com/view/constrained-manipulation/}. 

\end{appendices}

\end{document}